\pdfoutput=1
\documentclass[11pt]{article}
\newcommand{\miniprompt}[1]{
  \fbox{\mbox{#1}}
}
\newcommand{\added}[1]{#1}

\usepackage{algorithm}
\usepackage{algpseudocode}

\usepackage{adjustbox}
\usepackage{hyperref}

\usepackage{float}
\usepackage{stfloats}

\usepackage[preprint]{acl}

\usepackage{times}
\usepackage{latexsym}

\usepackage[T1]{fontenc}

\usepackage[utf8]{inputenc}

\usepackage{microtype}

\usepackage{inconsolata}

\usepackage{graphicx}

\usepackage{subcaption}
\usepackage{booktabs}
\usepackage{paralist}
\usepackage{amsmath}
\usepackage{amssymb}
\usepackage{todonotes}
\usepackage{cleveref}
\usepackage{CJKutf8} 
\newcommand{\zh}[1]{\begin{CJK}{UTF8}{gbsn}#1\end{CJK}}
\newcommand{\prompt}[1]{
\vspace{2mm}
\fbox{
\begin{minipage}{.85\linewidth}
\small
#1
\end{minipage}
}
\vspace{2mm}
}

\newcommand{\xhdr}[1]{\vspace{1.7mm}\noindent{{\bf #1.}}}
\newcommand{\cpt}[1]{\textsc{\MakeLowercase{#1}}}

\title{Separating Tongue from Thought: Activation Patching Reveals Language-Agnostic Concept Representations in Transformers}
\author{Clément Dumas\textsuperscript{12}\thanks{Equal contribution}\thanks{Work done while at EPFL} \
        Chris Wendler\textsuperscript{3}\footnotemark[1]\footnotemark[2] \\
        \textbf{Veniamin Veselovsky\textsuperscript{4}\footnotemark[2] \
        Giovanni Monea\textsuperscript{5}\footnotemark[2] \
        Robert West\textsuperscript{6}} \\
        \textsuperscript{1}ENS Paris-Saclay \
        \textsuperscript{2}Université Paris-Saclay \
        \textsuperscript{3}Northeastern \
        \textsuperscript{4} Princeton \
        \textsuperscript{5}Cornell \
        \textsuperscript{6}EPFL \\
        \texttt{\{clement.dumas@ens-paris-saclay.fr, chris.wendler@epfl.ch\}}}

\begin{document}
\maketitle

\begin{abstract}

A central question in multilingual language modeling is whether large language models (LLMs) develop a universal concept representation, disentangled from specific languages. In this paper, we address this question by analyzing latent representations (latents) during a word-translation task in transformer-based LLMs. We strategically extract latents from a source translation prompt and insert them into the forward pass on a target translation prompt. By doing so, we find that the output language is encoded in the latent at an earlier layer than the concept to be translated. Building on this insight, we conduct two key experiments. First, we demonstrate that we can change the concept without changing the language and vice versa through activation patching alone. Second, we show that patching with the mean representation of a concept across different languages does not affect the models' ability to translate it, but instead improves it. Finally, we generalize to multi-token generation and demonstrate that the model can generate natural language description of those mean representations.
Our results provide evidence for the existence of language-agnostic concept representations within the investigated models.\footnote{This work has been previously published under the title ``How Do Llamas Process Multilingual Text? A Latent Exploration through Activation Patching'' at the ICML 2024 Mechanistic Interpretability Workshop  \href{https://openreview.net/forum?id=0ku2hIm4BS}{https://openreview.net/forum?id=0ku2hIm4BS}.\\
Code for reproducing our experiments is available at \href{https://github.com/Butanium/llm-lang-agnostic}{https://github.com/Butanium/llm-lang-agnostic}}

\end{abstract}

\begin{figure}[ht]
    \centering
    \includegraphics[width=\linewidth]{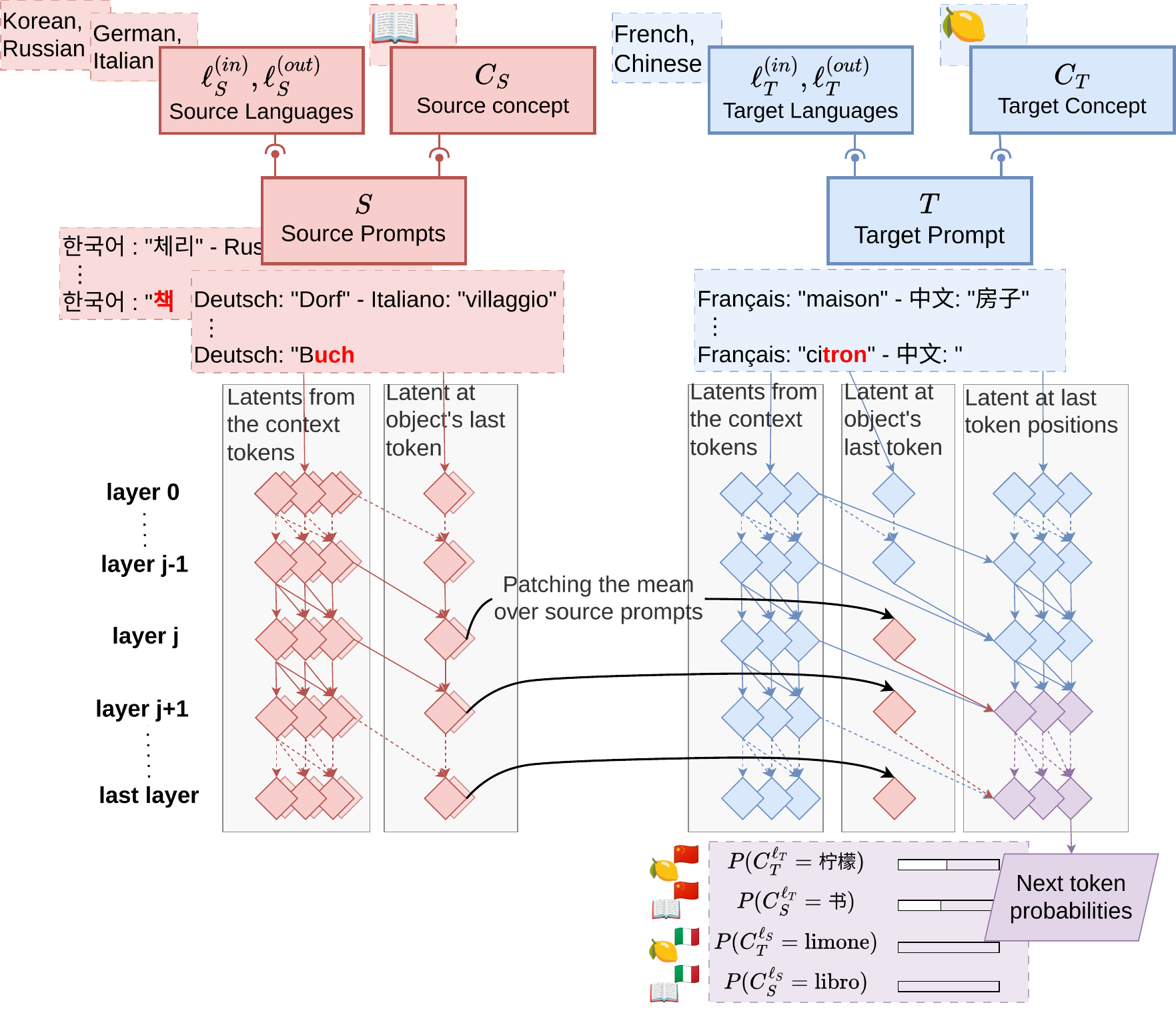}
    \caption{For two given concepts, e.g., \cpt{BOOK} and \cpt{LEMON}, we construct multiple source prompts which translate \cpt{BOOK}, and a target prompt for translating from French to Chinese. Then we extract the residual stream of the last token of the word to be translated after some layer $j$ and all subsequent ones from the source prompts and insert the mean of each layer at the corresponding positions in the forward pass of the target prompt. The resulting next token probabilities will concentrate on the \emph{source concept in target language} ($\text{\cpt{BOOK}}^\text{\cpt{ZH}}$, i.e., \zh{书}) when patching at layers 0--15, on the \emph{target concept in target language} ($\text{\cpt{LEMON}}^\text{\cpt{ZH}}$, \zh{柠檬}) for layers 16--31.}
    \label{fig:exp2}
    \vspace{-1.0em}
\end{figure}

\section{Introduction}
\label{sec:introduction}

\added{Most modern large language models (LLMs) are trained on massive corpora dominated by English text~\cite{ touvron2023llama, dubey2024llama, radford2019gpt2, brown2020gpt3, open2023gpt4}. Despite this imbalanced training, they achieve remarkable performance across multiple languages~\cite{shi2022language}, raising fundamental questions about how they process and represent multilingual information. Understanding these mechanisms is crucial not only to improve model performance, but also to identify potential biases and limitations in cross-lingual processing.}

\added{A fundamental question in multilingual language modeling is whether LLMs develop universal concept representations that transcend specific languages~\cite{wendler2024llamas, conneau2020emerging, chi2020findinguniversalgrammaticalrelations, xie-etal-2022-discovering, mousi2024exploring}. For example, when a model processes the word ``cat'' in English and ``chat'' in French, does it map these to the same internal representation of the feline concept, or does it maintain separate language-specific representations? Several recent works hint at the existence of language-agnostic concept representations.}

\added{\citet{wendler2024llamas} found that for simple multilingual tasks independent of the input and output language intermediate decodings of concept representations using the logit lens~\cite{nostalgebraist2020logitlens} decode to the English before they decode to the target language. Additionally, it has been long observed that instruction and safety tuning LLMs only on English data  generalizes to other languages~\cite{li2024preference, shaham2024multilingual}. The presence of language-agnostic representation in the pretrained LLMs would provide an explanation for both of these behaviors.}

This provides a unique opportunity for us to dig deeper and examine how multilingual concepts are represented and processed within LLMs, potentially revealing insights into language biases and concept formation. In particular, we are inspired by recent mechanistic interpretability approaches based on activation patching~\cite{variengien2023look, ghandeharioun2024patchscope, chen2024selfie}. These approaches are based on the idea of patching activations from one forward pass into another while observing the output (c.f.~Fig.~\ref{fig:1}) and present a simple, yet effective way to inspect the representations learned and causally understand the computations performed by LLMs.

\xhdr{Contributions} While prior work has provided observational evidence for shared semantic spaces in LLMs, we present the first causal analysis of how these representations are actively utilized during multilingual processing. More specifically, in this work we aim to understand how transformers process and represent concepts during translation tasks, whether language and concept information can be manipulated independently in the model's computations, and, whether models maintain separate language-specific concept representations or develop a shared conceptual space. To this end, we make the following contributions.

\begin{compactenum}
    \item First, we perform an activation patching analysis of Llama 2 7B~\cite{touvron2023llama}. We demonstrate that the model processes translation tasks by first resolving output language, then the concept to be translated.
    \item We propose two competing hypotheses about how transformers solve the translation task during their forward pass: \textbf{H1} where language and concepts are represented independently, and \textbf{H2} where they are inherently entangled. We argue that if language and concepts are independent (\textbf{H1}), averaging the latent representation of a concept across languages should still allow the model to make sense of and utilize this representation. Conversely, if language and concepts are entangled (\textbf{H2}), this mean representation would be an incoherent mixture of language-specific concepts that the model cannot effectively use.
    \item To test these hypotheses, we use a novel activation patching setup depicted in Figure~\ref{fig:exp2} which forces Llama 2 7B to translate this mean representation across languages. We find that using the mean concept representation across languages \emph{improves} Llama 2 7B's performance on a word translation task, supporting \textbf{H1}.
    \item We show that our observations generalize to a diverse set of transformer models varying in size, architecture, and training data, including Llama 2 70B, Llama 3 8B~\cite{dubey2024llama}, Mistral 7B~\cite{jiang2023mistral}, Qwen 1.5 7B~\cite{bai2023qwentechnicalreport}, Aya 23 8B~\cite{aryabumi2024aya23openweight} and Gemma 2 2B~\cite{gemmateam2024gemma2improvingopen}.
    \added{\item Finally, to support our claim that mean representation are usable by the model in a autoregressive generation setting, we present a novel activation patching setup depicted in Figure~\ref{fig:exp_gen} to show that a model can successfully write a definition of such a mean representation.
    }
\end{compactenum}

\xhdr{Implications} While prior work has suggested the existence of shared semantic spaces across languages through observational methods, our causal analysis provides the first direct evidence that LLMs actively utilize language-agnostic concept representations during text generation. Furthermore, our activation patching methodology establishes a framework for future causal investigations of multilingual representations, moving beyond the limitations of embedding similarity, probing, and logit lens approaches.

\section{Related Work}
\label{sec:related_work}
LLMs have demonstrated remarkable capabilities in processing multilingual text across languages, with examples including encoder-only model like mBERT \cite{devlin2018bert}, XLM-R \cite{conneau2020unsupervised}, and mT5 \cite{Xue_2021} and decoder-only model like \cite{aryabumi2024aya23openweight, dubey2024llama}. 
Studies on encoder-only models have shown that they develop language-agnostic representations, explaining their cross-lingual transfer capabilities. The methodology used was embedding similarity analysis \cite{conneau2020emerging, libovicky2020language, muller2021first, mousi2024exploring} and probing methods \cite{choenni2020does, pires2019multilingualmultilingualbert}.

While decoder-only transformers are not primarily designed to develop contextual embedding, but rather for open-ended text generation, they also develop cross-lingual generalization, for example, during safety and instruction tuning~\cite{li2024preference, chirkova2024zero}.
Mechanistic interpretability has led to powerful tools to analyze the language-agnosticity of these models. Using neuron analysis \citet{stanczak2022same, chen2024journey, cao2024one, zeng2024converginglinguafrancaevolution, tang2024language} have shown that LLMs develop both language-agnostic and language-specific neurons. \citet{wendler2024llamas, wu2024semantichubhypothesislanguage} use the logit lens~\cite{nostalgebraist2020logitlens} to perform early decoding during the forward pass of both LLMs and show that, no matter the language or input modality, the intermediate decodings concentrate on English before decoding to a specific language in the very last layers.

While the analyses via embedding similarity, probing and the logit lens give use valuable insight into the structure of the representation, they are not causal. Additionally, while the neuron level analysis studies the causal effects of the neurons, they do not study the representation themselves. Our work aims to fill this gap.

A related line of work has explored using definition generation as a means to evaluate semantic representations. \citet{noraset2017definition} introduced \emph{definition modeling}, the task of generating dictionary definitions from word embeddings, as a more direct evaluation of what semantic information embeddings capture. This approach has been extended to evaluate various types of embeddings \cite{gardner2022definitionion, chang2019does}. For example, \cite{chang2019does} demonstrated that contextualized embeddings like ELMo and BERT can be effectively mapped to definition spaces, revealing their sense-specific semantic content. In our work, instead of training a model to generate descriptions of LLM's representations, we repurpose the LLM itself to either translate or define it using activation patching. Activation patching, introduced by \citet{meng2022locating}, has been the main tool used to draw causal interpretation of LLM representations
~\cite{variengien2023look, geiger2022inducing, kramar2024atp*}. More recently, \citet{ghandeharioun2024patchscope, chen2024selfie} also show patching setups in which they can use the model itself to analyze its own model internal. Inspired by those methods, we developed two novel patching experiments supporting the language-agnosicity of LLMs representation.

Parallel work to ours by \citet{fierro2025multilingual} leverages the mechanistic interpretability toolkit to understand how factual recall works in multilingual LLMs as well. Similar to us, they find that the concept to be decoded and its language does not enter at the same layer into the residual stream, however in their knowledge association tasks they observe the opposite order than we do in our translation task. They also find that large parts of multilingual-factual recall are handled in a language agnostic way, despite the investigated models being trained on a more balanced split of languages than the ones studied in our paper.



\section{Background}
\label{sec:background}

\xhdr{Transformers forward pass}
When an autoregressive decoder-only transformer~\cite{vaswani2017attention, touvron2023llama} with $L$ layers processes a sequence of input tokens $x_1, \dots, x_n\in V$ from a vocabulary $V$, each token is initially transformed into a $d$-dimensional vector $h_i^{0}$ by an embedding layer. This first set of vector is the beginning of the residual stream. Then, for each token position $i$, the layer $j\in1,\dots, L$ updates the residual stream the following way:
 \begin{equation}
        h^{(j)}_i = h^{(j-1)}_i + f_{j}\left(h^{(j-1)}_1, \dots, h^{(j-1)}_{i}\right)
\end{equation}
where $f_j$ represents the operations of the $j$-th layer (typically self-attention followed by a feedforward network).
Finally, for a $m$-layer transformer, the next-token probabilities are obtained via a learned linear layer followed by a softmax operation mapping $h^{(m)}_i$ to $P(x_{i+1} | x_{1}\dots x_i)$.

\xhdr{\added{Activation patching}}\label{sec:actpatch}
\added{Activation patching is a causal intervention technique that allows us to understand how different components of a neural network contribute to its output. The key idea is to run two forward passes through the network -- one on a source input and one on a target input -- and copy (or ``patch'') activations from specific positions and layers of the source forward pass into the target forward pass. By observing how these interventions affect the model's output, we can understand what information is encoded in different parts of the network and how it is used.}

\added{More formally, given a source input $S$ and target input $T$, we can patch activations at position $i,i'$ and layer $j$ by setting $h_i^{(j)}(T) = h_{i'}^{(j)}(S)$ during the target forward pass, where $h_i^{(j)}$ represents the activation at position $i$ and layer $j$. The change in the model's output distribution provides evidence about what information was contained in the patched activation.}

\xhdr{Concepts}
We use capitalization to denote an abstract concept, (e.g. \cpt{CAT}). Let $C$ be an abstract concept, then we denote $C^{\ell}$ its language-specific version. Further, we define $w(C^{\ell})$ as the set of words expressing the abstract concept $C$ in language $\ell$. For example, if $C = \text{\cpt{CAT}}$ and $\ell = \text{\cpt{EN}}$ we have $w(C^{\ell}) =\ \{\text{``cat''}\}$ and similarly $w(C^{\text{\cpt{DE}}}) =\ \{\text{``Katze''}, \text{``Kater''}\}$.
Note that we talk about words for the sake of simplicity. However, on a technical level $w$ refers to the set of starting tokens of these words (e.g. $\{\text{``Katze''}, \text{``Kat''}\}$).
Therefore, when we track different sets of tokens $W$, (e.g. $W \in \{w(C_1^\cpt{IT}), w(C_1^{\cpt{ZH}}), w(C_2^{\cpt{IT}}), w(C_2^{\cpt{ZH}}), w(C_1^{\cpt{EN}}) \cup w(C_2^{\cpt{EN}})\} = \mathcal{W}$), we ensure that there is no token in common between any pair of $W_1, W_2 \in \mathcal{W}$ with $W_1 \neq W_2$.

\xhdr{Prompt design}\label{promptdesign}
We use the same translation prompt template as \cite{wendler2024llamas} that we denote TP(input language, output language, concept). For example, $\text{TP}(\cpt{EN}, \cpt{FR}, \cpt{CLOUD})$ could be:

\prompt{
English: ``computer'' - Français: ``ordinateur''\vspace{-0.3em}\\
...\\
English: ``ant'' - Français: ``fourmi''\\
English: ``cloud'' - Français: ``
}\label{prompt:TP}

\noindent{Here the task is to translate $w(\text{\cpt{CLOUD}}^\text{\cpt{EN}}) = \{\text{``cloud''}\}$ into $w(\text{\cpt{CLOUD}}^\text{\cpt{FR}}) = \{\text{``nuage''}\}$.}

More formally, for a given concept $C$, input language $\ell^{(\text{in})}$, and output language $\ell^{(\text{out})}$, we construct a few-shot translation prompt $\text{TP}(\ell^{(\text{in})}, \ell^{(\text{out})}, C)$. This prompt contains $k$ examples\footnote{in our study we used $k=5$} of single-word translations  of concepts $C_1,\dots,C_k$ from $\ell^{(\text{in})}$ to $\ell^{(\text{out})}$, concluding with the model being tasked to translate $C$ from $\ell^{(\text{in})}$ to $\ell^{(\text{out})}$. Using $C^{\ell^{(\text{in})}}$ as a shortcut for $v\in w(C^{\ell^{(\text{in})}})$, $\text{TP}(\ell^{(\text{in})}, \ell^{(\text{out})}, C)$ looks like:

\added{\prompt{
$\ell^{(\text{in})}$: ``$C_1^{\ell^{(\text{in})}}$'' - $\ell^{(\text{out})}$: ``$C_1^{\ell^{(\text{out})}}$''\vspace{-0.3em}\\
...\\
$\ell^{(\text{in})}$: ``$C_k^{\ell^{(\text{in})}}$'' - $\ell^{(\text{out})}$: ``$C_k^{\ell^{(\text{out})}}$''\\
$\ell^{(\text{in})}$: ``$C^{\ell^{(\text{in})}}$'' - $\ell^{(\text{out})}$: ``
}}

\noindent{We denote $\text{TP}^\text{concept}(\ell^{(\text{in})}, \ell^{(\text{out})}, C)$ as $\text{TP}(\ell^{(\text{in})}, \ell^{(\text{out})}, C)$ cut at the last token of $C^{\ell^{(\text{in})}}$.} For example, in our previous example, $\text{TP}^\text{concept}(\cpt{EN}, \cpt{FR}, \cpt{CLOUD})$ would be:

\prompt{
English: ``computer'' - Français: ``ordinateur''\vspace{-0.3em}\\
...\\
English: ``ant'' - Français: ``fourmi''\\
English: ``cloud
}

{\noindent We expect that the last token of such prompts is where the model stores its latent representation of $C^{\ell^{(\text{in})}}$.}

Importantly, whether the model correctly answers TP is determined by its next token prediction. In our \hyperref[prompt:TP]{prompt example}, the next token predicted should be ``nu'', the first token of ``nuage''.
Thus, we can track $P(C^{\ell})$\footnote{We use simplified notation $P(C^{\ell})$ rather than $P(C^{\ell}|\text{TP})$ throughout. While the conditional notation would be more precise for the initial case, our patching experiments involve multiple conditioning factors (target prompt, source prompt, patching configuration, layer) that would make the full notation unwieldy. We therefore adopt this simplified notation for clarity.}, i.e., the probability of the concept $C$ occurring in language $\ell$, by simply summing up the probabilities of all starting tokens of $w(C^{\ell})$ in the next-token distribution. 

We improve upon the construction of \citet{wendler2024llamas} by considering all the possible expressions of $C$ in $\ell$ using BabelNet~\cite{navigli2021babelnet}, instead of GPT-4 translations, when computing $P(C^{\ell})$. This allows us to capture many possible translations, instead of one. 
For example, ``commerce'', ``magasin'' and ``boutique'' are all valid words for $\text{\cpt{SHOP}}^\text{\cpt{FR}}$.

\section{Exploratory analysis with patching}
\label{sec:exploratory}

\xhdr{Problem statement} We aim to understand whether language and concept information can vary independently during Llama-2's forward pass when processing a multilingual prompt. For example, a representation of $C^{\ell}$ of the form $z_{C^{\ell}} = z_{C} + z_{\ell}$, in which $z_{C} \in U$, $z_{\ell} \in U^{\perp}$ and $U \oplus U^{\perp} = \mathbb{R}^{d}$ is a decomposition of $\mathbb{R}^d$ into a subspace $U$ and its orthogonal complement $U^{\perp}$, would allow for language and concept information to vary independently: language can be varied by changing $z_{\ell} \in U^{\perp}$ and concept by changing $z_{C} \in U$.
Conversely, if language and concept information were entangled, a decomposition like this should not exist: varying the language would mean varying the concept and vice versa.

\subsection{Experimental design} 
\begin{figure}[t]
    \centering
    \includegraphics[width=\linewidth]{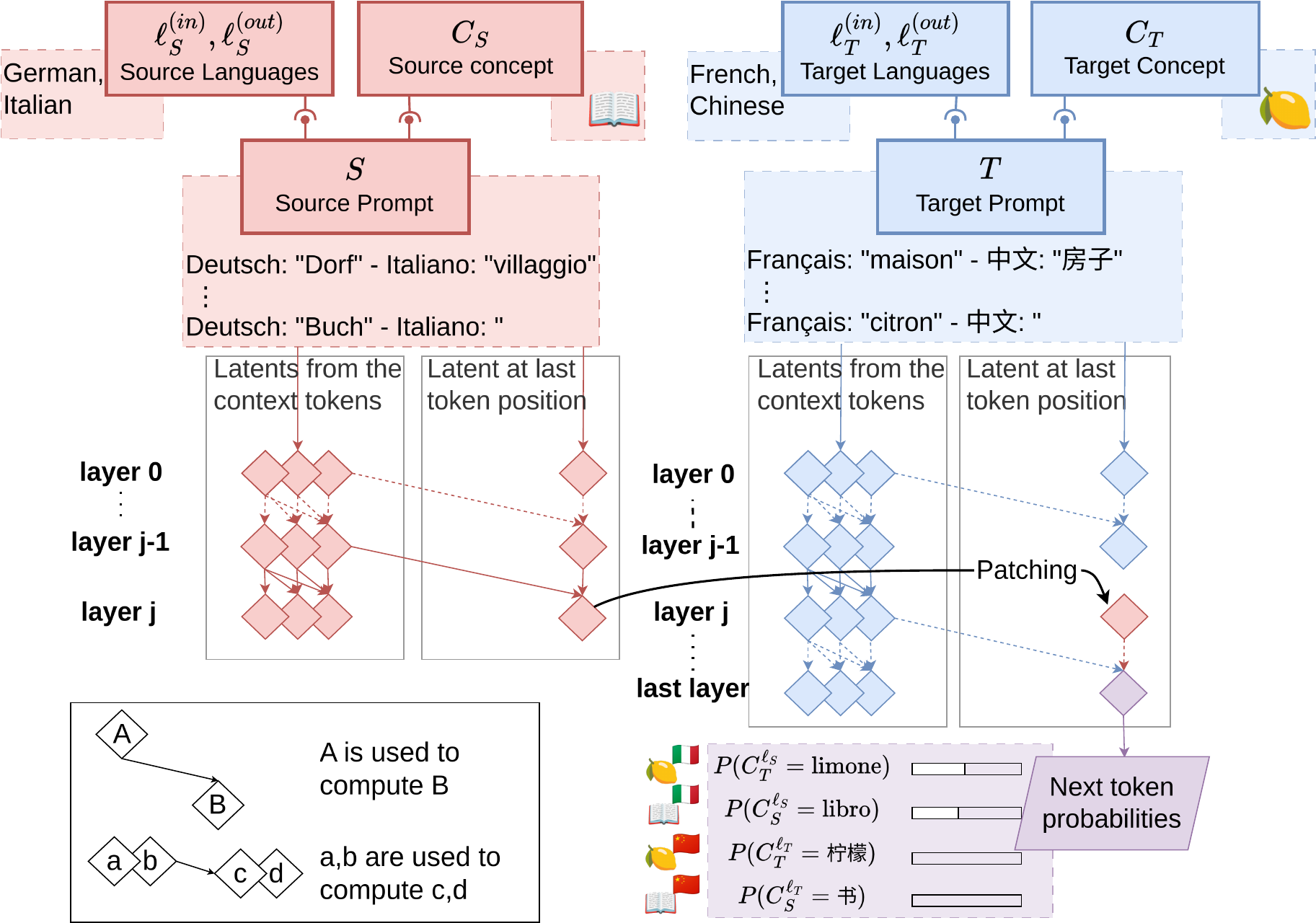}
    \caption{For two given concepts, e.g., \cpt{BOOK} and \cpt{LEMON}, we construct a source prompt for translating \cpt{BOOK} from German to Italian, and a target prompt for translating \cpt{LEMON} from French to Chinese. Then we extract the residual stream of the last token at a single layer $j$ from the source prompt and insert it at the corresponding position and layer in the forward pass of the target prompt. The resulting next token probabilities will concentrate on the \emph{target concept in the target language} ($\text{\cpt{LEMON}}^\text{\cpt{ZH}}$, i.e., \zh{柠檬}) when patching at layers 0--11, on the \emph{target concept in the source language} ($\text{\cpt{LEMON}}^\text{\cpt{IT}}$, ``limone'') for layers 12--16, and on the \emph{source concept in the source language} ($\text{\cpt{BOOK}}^\text{\cpt{IT}}$, ``libro'') for layers 17--31.}
    \label{fig:1}
\end{figure}

We start our analysis with an exploratory experiment on Llama 2 7B~\cite{touvron2023llama}. We use 5-shots translation prompts to create paired source $S=\text{TP}(\ell^{(\text{in})}_S, \ell^{(\text{out})}_S, C_S)$ and target prompt $T=\text{TP}(\ell^{(\text{in})}_T, \ell^{(\text{out})}_T, C_T)$ datasets with different concept, input languages and output languages\footnote{See details in \Cref{app:algopair}}.
If not mentioned otherwise, $\ell_{S}$ and $\ell_{T}$ refer to the output language of $S$ and $T$.

Similar to \cite{variengien2023look}, we would like to infer at which layers the output language and the concept enter the residual stream $h_{n_T}^{(j)}(T)$ respectively and whether they can vary independently for our task.
In order to investigate this question, we perform the experiment depicted in Figure~\ref{fig:1}.
For each transformer block $f_{j}$ we create two parallel forward passes, one processing the source prompt $S$ which tokens are $(s_1, \dots, s_{n_S})$ and one processing the target prompt $T = (t_1, \dots, t_{n_T})$. While doing so, we extract the residual stream of the last token of the source prompt after layer $j$, $h_{n_S}^{(j)}(S)$, and insert it at the same layer at position $n_T$  in the forward pass of the target prompt, i.e., by setting $h_{n_T}^{(j)}(T) = h_{n_S}^{(j)}(S)$ and subsequently completing the altered forward pass. From the resulting next token distribution, we compute $P(C_S^{\ell_S}), P(C_S^{\ell_T}), P(C_T^{\ell_S}), \text{ and } P(C_T^{\ell_T})$. 

\subsection{Results}

\begin{figure}[ht]
        \centering   
        \includegraphics[width=0.7\linewidth]{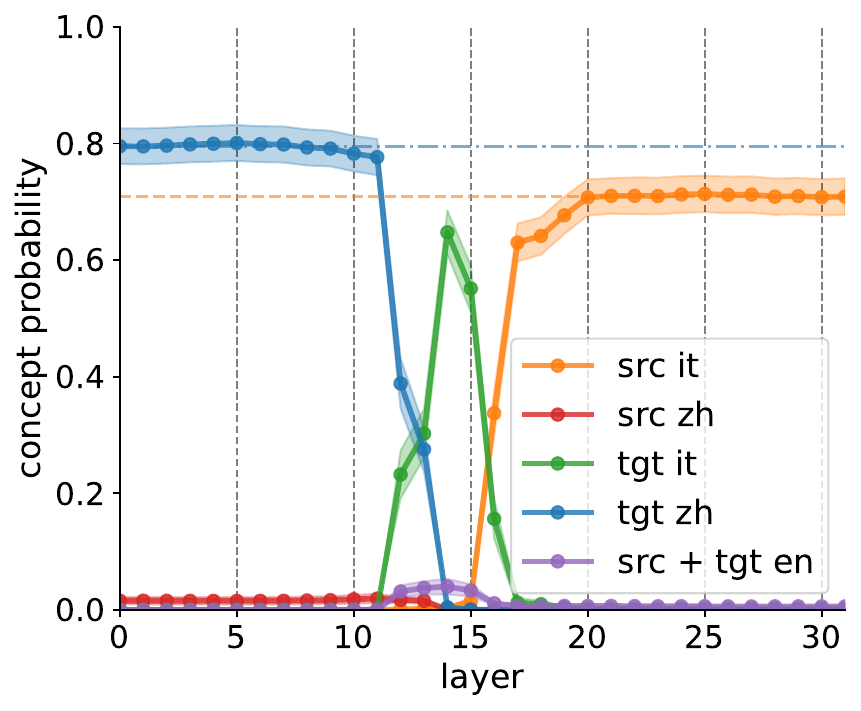}
     \caption{Our first patching experiment with a \cpt{DE} to \cpt{IT} source prompt and a \cpt{FR} to \cpt{ZH} target prompt with different concepts. The x-axis shows at which layer the patching was performed and the y-axis shows the probability of predicting the correct concept in language $\ell$ (see legend). In the legend, the prefix ``src'' stands for source and ``tgt'' for target concept. The orange dashed line and blue dash-dotted line correspond to the mean accuracy on source and target prompt. We report means and 95\% Gaussian confidence intervals computed over 200 source, target prompt pairs featuring 41 source concepts and 38 target concepts.}
     \label{fig:actpatching}
     \vspace{-1.0em}
 \end{figure}

In this experiment, we perform the patching at one layer at a time and report the probability that is assigned to $P(C_S^{\ell_S}), P(C_S^{\ell_T}), P(C_T^{\ell_S}), \text{ and } P(C_T^{\ell_T})$. 
As a result we obtain Figure~\ref{fig:actpatching} in which we report means and 95\% confidence interval over 200 examples.

\xhdr{Interpretation} We observe the following pattern while patching at different layers (see Figure~\ref{fig:actpatching}):
\begin{compactitem}
    \item Layers 0--11: Target concept decoded in target language, resulting in large $P(C_T^{\text{\cpt{ZH}}})$.
    \item Layers 12--16: Target concept decoded in source language, resulting in large $P(C_T^{\text{\cpt{IT}}})$.
    \item Layers 16--31: Source concept decoded in source language, resulting in large $P(C_S^{\text{\cpt{IT}}})$.
\end{compactitem}

This pattern suggests that the model first computes the output language: from layer 12 onwards, we decode in the source output language. This indicates that up until that layer, the need to decode to $\ell^{(out)}$ is being encoded in the residual stream and subsequently remains unchanged. For example, this could be achieved by the model computing a function vector $z_{\ell^{(out)}}$~\cite{todd2023function}. If this hypothesis is correct, patching at layer 12 would overwrite $z_{\ell_T^{(out)}}$ with $z_{\ell_S^{(out)}}$.
The green spike between layer 12 and 16 indicates that at those layer, the concept is not yet represented, so the model keep outputing the target concept but in the source language. 

In later layers, the model determines the concept: from layer 16 onwards, the source concept is decoded. This suggests that $z_{C_T^{\ell^{(\text{out})}}}$ is overwritten at layer 16.\footnote{In Appendix~\ref{app:right}, we collected additional experimental results investigating the right part of Figure~\ref{fig:actpatching} more deeply and in Appendix~\ref{app:left} the left part. For the right part, we use the patchscope lens~\cite{ghandeharioun2024patchscope} to investigate from which layer it is possible to decode the source concept in App.~Figure~\ref{fig:patchscope1_appendix}. The results of both experiments agree: from layer 16 it is possible to decode the source concept in source language. For the left part, we experiment with randomized source prompts and different prompting templates in between source and target prompt in App.~Figure~\ref{fig:randomandact}. We find that indeed before layer 11 there is no translation task specific information in the residual stream, only prompt-template specific information.}

\xhdr{Hypotheses} We are left with two hypotheses compatible with these results:
\begin{compactitem}
    \item \textbf{H1}: Concept and language are represented independently. When doing the translation, the model first computes ${\ell^{(\text{out})}}$ from context, and then identifies $C$. In the last layers, it then maps $C$ to the first token of $w(C^{{\ell^{(\text{out})}}})$.
    \item \textbf{H2}: The representation of a concept is always entangled with its language. When doing the translation, the model first computes ${\ell^{(\text{out})}}$, then computes $\ell^{(\text{in})}$ and $C^{\ell^{(\text{in})}}$ from its context and solves the language-pair-specific translation task of mapping $C^{\ell^{(\text{in})}}$ to $C^{\ell^{(\text{out})}}$.
\end{compactitem}


\section{Ruling out hypotheses}
\label{sec:ruling-out}

Next, we run additional experiments to (1) provide further evidence that we are either in \textbf{H1} or \textbf{H2} and (2) to disambiguate whether we are in \textbf{H1} or \textbf{H2} (3) to show that our findings hold for other models.

\xhdr{Further evidence experiment}
In the experiments in Sec.~\ref{sec:exploratory} we did not observe \emph{source concept in target language}. However, both \textbf{H1} and \textbf{H2} would allow for that to happen via patching in the right way. Therefore, in this experiment, instead of overwriting the residual stream at the last token of the prompt, we overwrite them at the last token of the word to be translated. \added{In order to do that, for the source prompt, we use $\text{TP}^\text{concept}$ instead of TP ($S = \text{TP}^\text{concept}(\ell^{(\text{in})}_S, \ell^{(\text{out})}_S, C_S)$).
This means that we collect the activations at the last token of $C_S^{\ell_S^{(\text{in})}}$.}

Let $\rho_T$ denote the position of that token in the target prompt. Since the concept information seems to enter via multiple layers (16-20) into the latent of the last token, we overwrite the latent corresponding to the token at position $\rho_T$ at layer $j$ \emph{and all subsequent ones}.
By patching in this way in both \textbf{H1} and \textbf{H2} we would expect to see large $P(C_S^{\ell_T})$.

Formally, we patch by setting $h_{\rho_T}^{(j)}(T) = h_{-1}^{(j)}(S), \dots, h_{\rho_T}^{(L)}(T) = h_{\added{-1}}^{(L)}(S)$\footnote{Note that we use Python indexing, where -1 denotes the last token.}.

\xhdr{Disambiguation experiment}
Both \textbf{H1} and \textbf{H2} compute $w(C_S^{\ell_{T}})$ but in different ways. In \textbf{H1} one decoding circuit per output language is required in order to compute the expression for the concept $C_S$ in language $\ell_T$. In contrast, in \textbf{H2} one translation circuit per input-output language pair is required to map the entangled $C^{\ell^{(\text{in})}_S}_S$ to $C^{\ell^{(\text{out})}_T}_S$. Therefore, in order to disambiguate the two, we construct a patching experiment that should work under \textbf{H1}, but fail under \textbf{H2}.

In order to do so, instead of patching the latent containing $C^{\ell_S^{(\text{in})}}_S$ from a single source forward pass, we create multiple source prompts with the same concept $C_S$ but in different input languages $\ell^{(\text{in})}_{S_1} \neq \dots \neq \ell^{(\text{in})}_{S_k}$ and output languages $\ell^{(\text{out})}_{S_1} \neq \dots \neq \ell^{(\text{out})}_{S_k}$ and patch by setting 
$$h_{\rho_T}^{(\alpha)}(T) = \frac{1}{k} \sum_{i = 1}^k h_{-1}^{(\alpha)}(S_i),$$
for $\alpha\in j,\dots, m$.
Let $C_i = C^{{\ell^{(\text{in})}_{S_i}}}_{S}$, under \textbf{H1}, taking the mean of several language-specific concept representations should keep the concept information intact, since for all $i$, $z_{C_i} = z_{C_S}$:
$$\frac{1}{k} \sum_{i = 1}^k z_{C_i} = 
z_{C_S} + \frac{1}{k} \sum_{i = 1}^k z_{\ell^{(\text{in})}_{S_i}}.$$
Therefore, we'd expect high $P(C_S^{\ell_T})$ in this case. However, under \textbf{H2}, in which $z_{C_i}$ cannot be disentangled, this mean representation may not correspond to a well-defined concept. Additionally, the interference between multiple input languages should cause difficulties for the language-pair-specific translation, which should result in a drop in $P(C_S^{\ell_T})$.



\xhdr{Results}
In the first experiment, we use the same languages as above and in the second one we used \cpt{DE}, \cpt{NL}, \cpt{ZH}, \cpt{ES}, \cpt{RU} as input and \cpt{IT}, \cpt{FI}, \cpt{ES}, \cpt{RU}, \cpt{KO} as output languages for the source prompts and \cpt{FR} to \cpt{ZH} for the target prompt.

\begin{figure}[ht]
    \begin{subfigure}[b]{0.49\linewidth}
        \centering 
        \caption{Single source prompt}
        \label{fig:obj}
        \includegraphics[width=\textwidth]{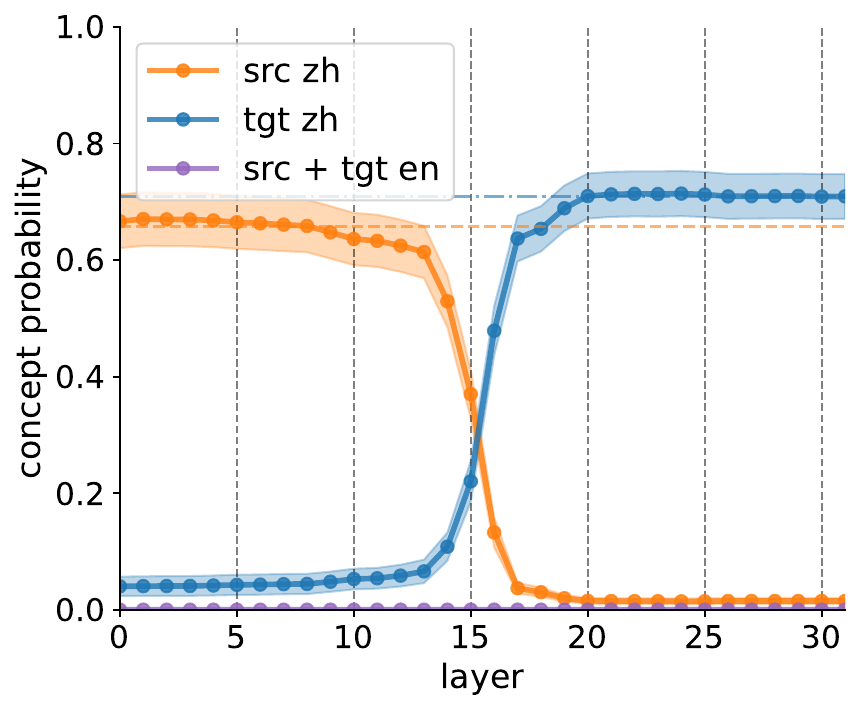}
        
    \end{subfigure}
    \hfill
     \begin{subfigure}[b]{0.49\linewidth}
         \caption{Mean over source prompts}
          \label{fig:mean_obj}
      \centering \includegraphics[width=\textwidth]{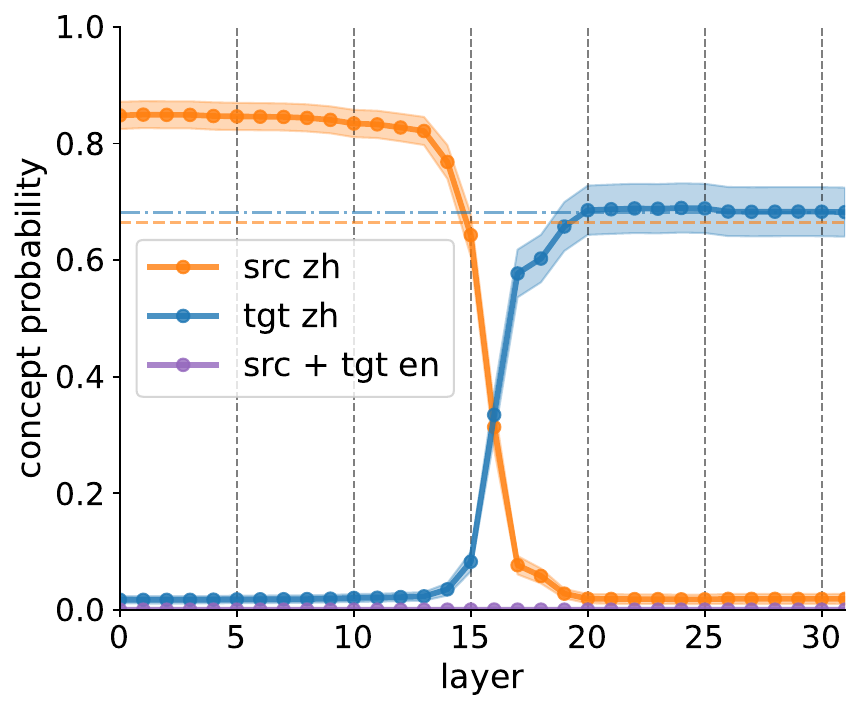}
    \end{subfigure}
    \caption{Here we use different input languages (\cpt{DE}, \cpt{FR}), different concepts, different output languages (\cpt{IT}, \cpt{ZH}) in (a). In (b) we use multiple source input languages \cpt{DE}, \cpt{NL}, \cpt{ZH}, \cpt{ES}, \cpt{RU} and source output languages \cpt{IT}, \cpt{FI}, \cpt{ES}, \cpt{RU}, \cpt{KO}. We patch at the last token of the concept-word at all layers from $j$ to $31$. In (a) we patch latents from the single source prompt in (b) we patch the mean of the latents over the source prompts. 
    For each of the plots, the x-axis shows at which layer the patching was performed during the forward pass on the target prompt and the y-axis shows the probability of predicting the correct concept in language $\ell$ (see legend). The prefix ``src'' stands for source and ``tgt'' for target concept. We report means and 95\% Gaussian confidence intervals computed over a dataset of size 200.}
    \label{fig:patchscope2}
 \end{figure}

In Figure~\ref{fig:patchscope2} we observe that in both experiments, we obtain very high probability for the \emph{source concept in the target language} $P(C_S^{\cpt{ZH}})$ from layers 0 to 15, i.e., exactly until the latents at the last token stop attending to the last concept-token. 

Therefore, Figure~\ref{fig:patchscope2}~(a) supports that we are indeed either in \textbf{H1} or \textbf{H2}, since \emph{as planned} we successfully decode \emph{source concepts in the target language} $P(C_S^{\cpt{ZH}})$ from layers 0 to 15. Conversely, if we were not able to decode \emph{source concept in target language} in this way this would have spoken against both \textbf{H1} and \textbf{H2}. 

Additionally, Figure~\ref{fig:patchscope2}~(b) supports that we are in \textbf{H1} and not in \textbf{H2} because patching in the mean keeps $P(C_S^{\cpt{ZH}})$ intact and even increases it. Therefore, instead of observing interference between the different language-entangled concepts as would have been predicted by \textbf{H2}, we observe a concept-denoising effect by averaging multiple language-agnostic concept representations which only makes sense under \textbf{H1}. Taking the mean over concept representations corresponding to different input languages seems to act like a majority voting mechanism resulting in an increase in $P(C_S^{\cpt{ZH}})$. \footnote{Conversely, e.g., averaging over different translation prompt contexts but while keeping the input and output language fixed does not lead to an increase in $P(C_S^{\cpt{ZH}})$ (see App.~Figure~\ref{fig:obj_patching_full},\ref{fig:obj_patching_full2}~(b)).}

\xhdr{Other models} In Appendix~\ref{app:other-models} we perform the experiments from Sec.~\ref{sec:exploratory} and Sec.~\ref{sec:ruling-out} on several other models, varying in size, training data and architecture namely, Mistral 7B~\cite{jiang2023mistral}, Llama 3 8B~\cite{dubey2024llama}, Qwen 1.5 7B~\cite{bai2023qwentechnicalreport}, Llama 2 70B, and Aya 23 8B~\cite{aryabumi2024aya23openweight} which was specifically trained to be multilingual. We observe the same improvement when we take the mean of a concept across languages for all these models, suggesting that they are all operating under H1 and use some language-agnostic concept representation.

\section{Generating descriptions for latents}
We just showed that LLMs can effectively translate concept representations averaged across multiple languages, providing evidence for language-agnostic concept representations. In this section, we explore whether these mean representations, which theoretically capture language-independent concepts, can be described by the model in natural language as effectively as concepts expressed in a single language. This approach builds on the definition modeling paradigm \cite{noraset2017definition, mickus2022semeval, chang2019does, gardner2022definitionion}, which uses natural language generation as a transparent evaluation of semantic representations.

\subsection{Definition prompt}
\added{In order to do that, we introduce a new prompt template that tasks the model to describe a concept in natural language. More precisely, given a concept $C$ and a language $\ell$, we construct a few shot definition prompt $\text{DP}(\ell, C)$ of the form:}

\added{\prompt{
    ``$C^{\ell}_1$'' : ``$D^{\ell}_1$''\vspace{-0.3em}\\
    ...\\
    ``$C^{\ell}$'' : ``
}}

\noindent where $C^{\ell}_1, \dots, C^{\ell}_n$ are concepts in language $\ell$ and $D^{\ell}_1, \dots, D^{\ell}_n$ are their descriptions in language $\ell$. We denote $\text{DP}^\text{concept}(\ell, C)$ the prompt template that ends at the last token of $C^{\ell}$.
For example, $\text{DP}^\text{concept}(\cpt{EN}, \cpt{ANT})$ could be:

\added{\prompt{
    ``apple'' : ``Fruit with red or yellow or green skin''\vspace{-0.3em}\\
    ...\\
    ``ant
}}

For each language, we constructed a definition dataset using the first concept in BabelNet~\cite{navigli2021babelnet} associated with each of the 200 picturable words from the Basic English word list from Wikipedia\footnote{\url{https://en.wiktionary.org/wiki/Appendix:Basic_English_word_list\#Things_-_200_picturable_words}}. For each concept, BabelNet provides several definitions in different languages.

\subsection{Patching setup}
\added{For languages $\ell_S^1, \dots, \ell_S^n$ and $\ell_T$, and concepts $C_S \neq C_T$, we construct a target prompt $T=\text{DP}(\ell_T, C_T)$ and two sets $\mathcal{S}$ of source prompts:}
\begin{compactitem}
    \item \textbf{From translations}: for each language $\ell_S^i$ we pick an input language $\ell_{(\text{in})}^i$ and choose $$\mathcal{S}_\text{trans} = \{\text{TP}^\text{concept}(\ell_{(\text{in})}^i, \ell_S^i, C_S)\}_{i \in \{1, \dots, n\}}.$$
    \item \textbf{From definitions}: we choose $$\mathcal{S}_\text{def} = \{\text{DP}^\text{concept}(\ell_S^i, C_S)\}_{i \in \{1, \dots, n\}}.$$
\end{compactitem}

\begin{figure}[ht]
    \centering
    \includegraphics[width=\linewidth]{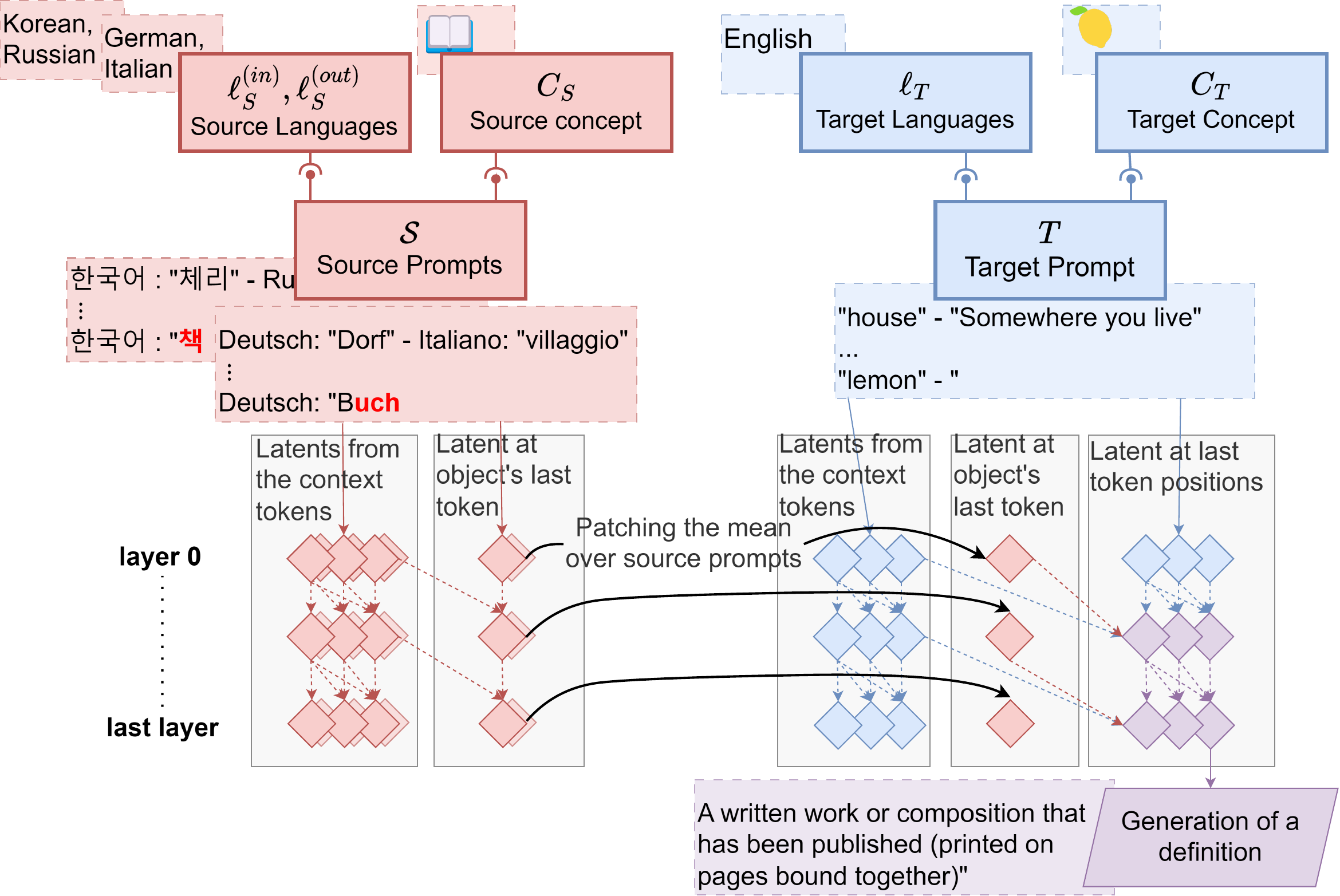}
    \caption{\added{Illustration of the patching setup for the definition prompt experiment. We patch the latents of the last token of the source prompts from $\mathcal{S}$ to the last token of $C_T$ in the target prompt.}}
    \label{fig:exp_gen}
\end{figure}

\added{Then, to generate a definition of $C_S$ in language $\ell_T$, for all layers, we patch the latents of the last token averaged over the source prompts from $\mathcal{S}$ to the last token of $C_T$ in the target prompt and let the model generate the definition as depicted in Figure~\ref{fig:exp_gen}. More formally, we patch by setting $$h_{\rho_T}^{(j)}(T) = \frac{1}{n} \sum_{i = 1}^n h_{-1}^{(j)}(S_i)$$ for $j\in\{1, \dots, m\}$ and $\rho_T$ the position of the last token of $C_T$ in the target prompt.}

\added{\subsection{Experiment}}
\added{To compare the quality of the definitions of $C_S$ generated by the model in our experiment, we use the following baselines:}
\added{\begin{compactitem}
    \item \textbf{Ground truth}: We use a random definition from BabelNet.
    \item \textbf{Prompting}: We use the definition generated by the model when prompted with $\text{DP}(\ell_T, C_S)$.
    \item \textbf{Word patching}: We replace $C_T^{\ell_T}$ with $C_S^{\ell_S^j}$ for a random $j\in\{1, \dots, n\}$ and let the model generate the definition. In this setting, the model is tasked to generate a definition of a word in a language different from the target prompt's language.
    \item \textbf{Repeat word}: We use $C_S^{\ell_T}$ as the definition.
\end{compactitem}}

\added{To evaluate the different definitions, we use the \texttt{sentence-transformers} library~\cite{reimers-2019-sentence-bert} to run \texttt{paraphrase-multilingual-mpnet-base-v2}\footnote{\url{https://huggingface.co/sentence-transformers/paraphrase-multilingual-mpnet-base-v2}}, a semantic similarity model distilled from \cite{song2020mpnetmaskedpermutedpretraining} using the methods from \cite{reimers-2020-multilingual-sentence-bert}. To assess the quality of the generated definition, we compute the similarity score between embeddings of the definition and the mean embedding of the ground truth definitions\footnote{To be able to compare the score of the generated definition with the ground truth baseline, we compute its similarity score with the mean embedding of the \textbf{other} ground truth definitions - excluding the one that was randomly chosen as the ground truth baseline}.}

\added{\subsection{Results}}
\begin{figure*}[t]
    \centering
    \includegraphics[width=\linewidth]{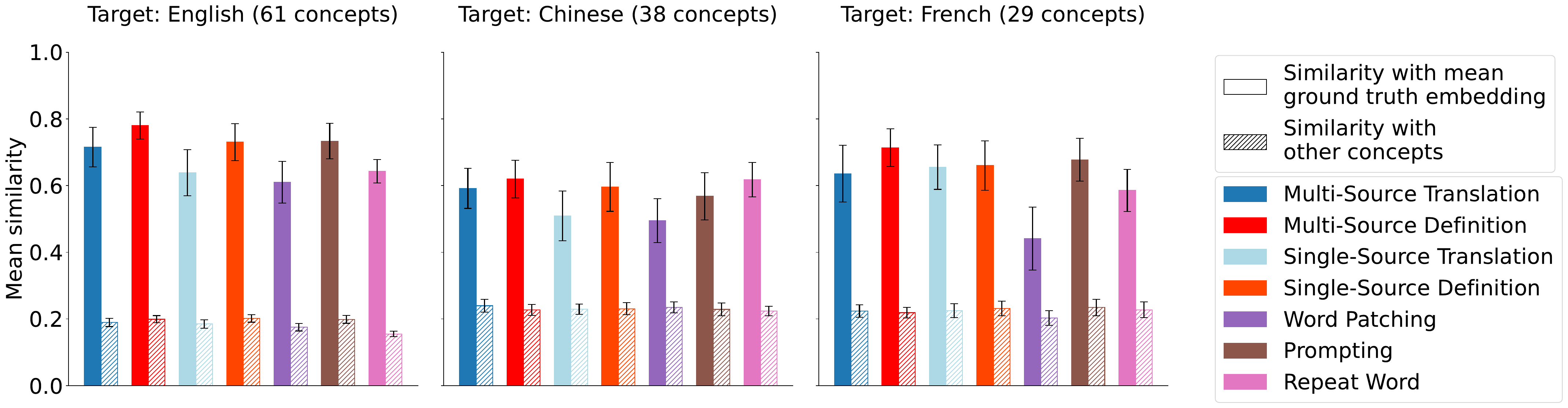}
    \caption{Mean similarity between the definition and the mean embedding of the ground truth definitions, as well as the mean similarity between the definition embedding and the embeddings of the definitions of the other concepts in the dataset. The results are presented for three target languages: English (with source languages Italian, Finnish, Spanish, Korean, and input languages for source translations: German, Dutch, Chinese, Russian), Chinese (using the same languages), and French (with source languages Korean, Japanese, Estonian, Finnish and English as input language for source translations).
    We report means and 95\% Gaussian confidence intervals computed over a dataset of various sizes\footnotemark.}
    \label{fig:def_sims}
\end{figure*}
\footnotetext{For some concepts and languages, BabelNet does not provide any definition.}
We report the mean similarity score between the definition and the mean embedding of the ground truth definitions for LLama 2 7B in Figure~\ref{fig:def_sims}. To give an idea of the scale of the similarity scores, we also report the mean similarity between the definition embedding and the embeddings of the definitions of the other concepts in the dataset.

\added{We find that patching concept representations from one language to another allows the model to generate high-quality definitions, comparable to or better than direct prompting. The fact that patching mean representations across multiple source languages leads to slightly better results suggests that the model's concept representations are indeed language-agnostic -- if they were language-specific, averaging across languages would likely degrade performance. This aligns with our translation experiment findings and adds another perspective on how LLMs process multilingual information.}

\added{Additionally, the comparable performance between patching from definitions and translations indicates that the model builds similar concept representations regardless of whether it processes a translation or definition prompt. This suggests a unified internal representation of concepts that generalizes across different types of language tasks.}

\added{\xhdr{Other models} In Appendix~\ref{app:similarities} we show that those findings generalize to other models and languages.}
\section{Conclusion}

In this paper, we showed that transformers use language-agnostic latent representations of concepts when processing word-level translation prompts. We achieved this by patching latents between parallel forward passes for translation prompts that differed in both input and output languages, as well as in the specific concepts being translated. Our main finding was that translation performance \emph{improves} when the transformer is forced to translate a concept representation averaged across multiple languages. This finding speaks for language-agnostic concept representations. As we argued, for language-agnostic concept representations, taking the mean representation of a concept across languages should not impair the LLM's ability to translate this concept. In contrast, for language-specific ones, taking the mean should result in interference between the different language-specific versions of the concept. Thus, our results are consistent with findings from previous work \cite{wendler2024llamas} indicating that Llama-2 represents concepts in a concept space independent of the language of the prompt. Our work also provides evidence that findings from BERT models \cite{conneau2020emerging, pires2019multilingualmultilingualbert} generalize to a wide range of decoder-only transformers. Our findings open several important avenues for future research. Understanding these disentangled representations could improve cross-lingual transfer learning and reduce the computational costs of multilingual training by leveraging shared concept spaces more efficiently. Moreover, our results provide mechanistic insights into the root causes of Western cultural biases in multilingual LLMs, suggesting that biases may propagate through the shared concept space. We hope these contributions will guide efforts to build more controllable, efficient, and culturally-aware multilingual language models.
\section*{Limitations}

In this work, we studied how transformers represent concepts when processing multilingual text. \added{However, we only considered very simple concepts, maybe some more complex concepts would have shown a different behavior. Also, we did not study language-specific concepts like ``Waldeinsamkeit'', ``The feeling of solitude and connectedness to nature when being alone in the woods.''. It would be interesting to see how those are represented.}

Furthermore, more fine-grained probing will be required to determine to which degree transformers are able to specialize a concept to a language and whether concepts and languages are entangled in more subtle ways.
    
\section*{Acknowledgment}
We would like to thank the team working on \texttt{NNsight}~\cite{nnsight} which is the python package we used to implement all our experiments. We thank Hannes Wendler for multiple fruitful discussions.

\bibliography{biblio}

\appendix

\section{Patchscope experiment}
\label{app:right}

\begin{figure}[ht]
    
    \begin{subfigure}[t]{0.49\linewidth}
        \centering   
        \caption{Activation patching}
        \label{fig:actpatching_appendix}
        \includegraphics[width=\linewidth]{figures/plots/every_diff2.pdf}
    \end{subfigure}
    \hfill
    \begin{subfigure}[t]{0.49\linewidth}
    \caption{Patchscope lens}
    
    \includegraphics[width=\linewidth]{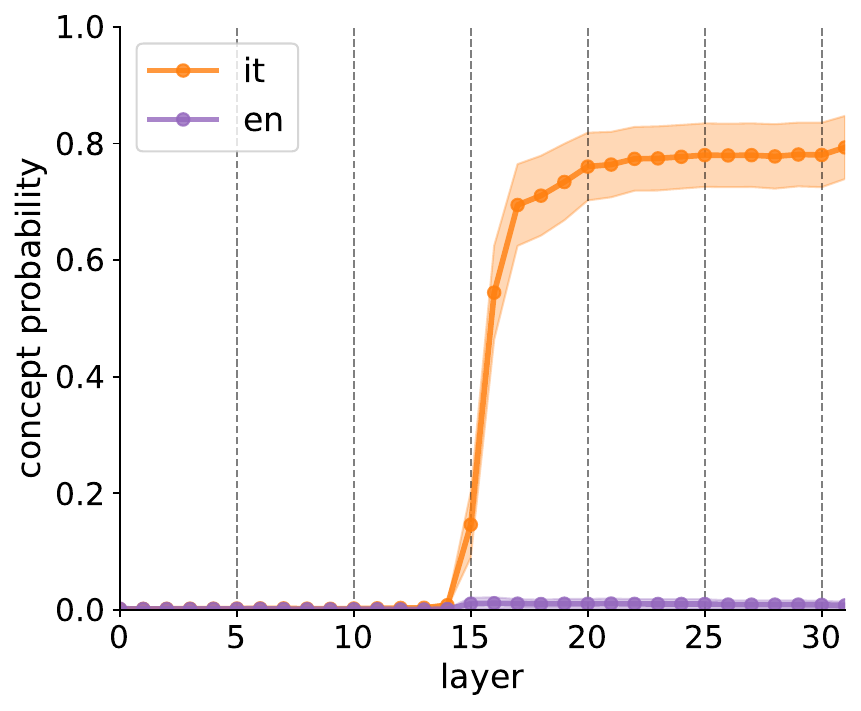}
    \end{subfigure}
     \caption{(a) Our first patching experiment with a \cpt{DE} to \cpt{IT} source prompt and a \cpt{FR} to \cpt{ZH} target prompt with different concepts. (b)~Our patchscope lens experiment with a \cpt{DE} to \cpt{IT} source prompt and identity target prompt \miniprompt{king king\textbackslash n1135 1135\textbackslash nhello hello\textbackslash n?}. We patch at the last token respectively. For each of the plots the x-axis shows at which layer the patching was performed during the forward pass on the target prompt and the y-axis shows the probability of predicting the correct concept in language $\ell$ (see legend). In the legend the prefix ``src'' stands for source and ``tgt'' for target concept. The orange dashed line and blue dash-dotted line correspond to the mean accuracy on source and target prompt. We report means and 95\% Gaussian confidence intervals computed over 200 source-, target prompt pairs featuring 41 source concepts and 38 target concepts for (a) and 38 prompts for (b).}
     \label{fig:patchscope1_appendix}
     \vspace{-1.0em}
 \end{figure}

We performed an additional experiment using the patchscope lens~\cite{ghandeharioun2024patchscope} to collect more evidence about from which layer it is possible to decode the source concept in Figure~\ref{fig:patchscope1_appendix}. The results of this experiment corroborate the findings presented in Section~\ref{sec:exploratory}. To enable a convenient comparison of the experimental results, we also include Figure~\ref{fig:actpatching} in Figure~\ref{fig:patchscope1_appendix}.

\section{Translation Pair Construction}
\label{app:algopair}

To ensure reproducibility of our experiments, we provide the pseudocode for constructing translation pairs used in our activation patching experiments. The complete implementation is available in our codebase at \texttt{notebooks/obj\_patch\_translation.ipynb}.
\begin{algorithm}
\caption{Construction of Translation Pairs for Patching Experiments}
\label{alg:translation-pairs}
\begin{algorithmic}[1]
\Require Set of source languages $\mathcal{L}_S$, target language $\ell_T$, number of pairs $n$
\Ensure Set of valid translation pairs $\mathcal{P}$
\State Load BabelNet translations for all languages
\State $\mathcal{T}_S \gets$ \Call{GetTranslations}{$\mathcal{L}_S$}
\State $\mathcal{T}_T \gets$ \Call{GetTranslations}{$\ell_T$}
\State $\mathcal{P} \gets \emptyset$ \Comment{Initialize valid pairs}
\State $\mathcal{A} \gets$ \Call{Combinations}{$\mathcal{T}_S, \mathcal{T}_T$} \Comment{All possible pairs}
\State \Call{Shuffle}{$\mathcal{A}$} \Comment{Randomize order}
\For{$(w_S, w_T) \in \mathcal{A}$}
    \If{$\text{concept}(w_S) = \text{concept}(w_T)$}
        \State \textbf{continue} \Comment{Skip same concepts}
    \EndIf
    \If{\Call{HasTokenCollisions}{$w_S, w_T$}}
        \State \textbf{continue} \Comment{Skip pairs with token overlap}
    \EndIf
    \State $\mathcal{P} \gets \mathcal{P} \cup \{(w_S, w_T)\}$
    \If{$|\mathcal{P}| \geq n$}
        \State \textbf{break} \Comment{Sufficient pairs collected}
    \EndIf
\EndFor
\State \Return $\mathcal{P}$
\end{algorithmic}
\end{algorithm}

\noindent\textbf{Key constraints:}
\begin{itemize}
    \item \textbf{Different concepts}: We ensure that source and target words represent different concepts to enable meaningful patching experiments.
    \item \textbf{No token collisions}: As described in Section~\ref{sec:background}, we track sets of tokens $w(C^{\ell})$ for each concept-language pair. To ensure clean probability measurements, we verify that there is no overlap between the token sets of paired concepts across all languages used in the experiment.
    \item \textbf{Randomization}: Pairs are shuffled before selection to avoid systematic biases in concept or language selection.
\end{itemize}

For our experiments, we typically use $n=200$ pairs, with concepts drawn from the 200 picturable words from the Basic English word list, ensuring sufficient statistical power while maintaining computational feasibility.

\section{Random prompt task experiment}
\label{app:left}

In order to investigate the leftmost part of Figure~\ref{fig:actpatching_appendix} more deeply, we performed additional experiments in which we explore ``random'' source prompts instead of translation source prompts.

The experimental setting here is similar to the one in Sec.~\ref{sec:exploratory},
except for the fact that instead of patching in latents from a translation source prompt we patch latents from different ``random'' source prompts. For the random source prompts, we gradually move away from the prompting template.

\begin{figure}[ht]
    \centering
    \begin{subfigure}[t]{0.49\linewidth}
        \caption{Random prompt}
        \label{fig:random}
        
        \includegraphics[width=\linewidth]{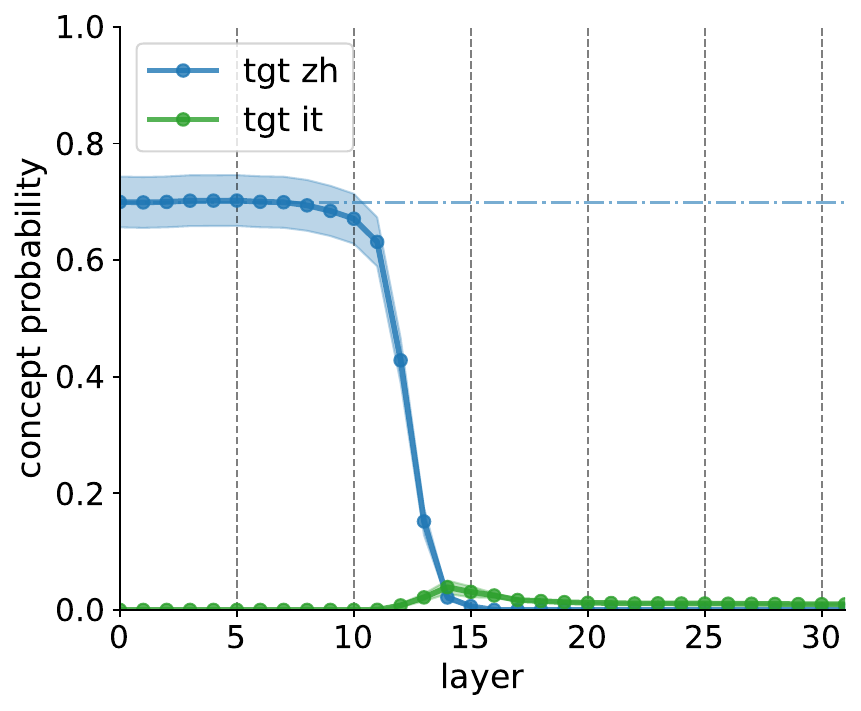}
    \end{subfigure}
    \hfill
    \begin{subfigure}[t]{0.49\linewidth}
        \caption{Empty prompt}
        \label{fig:random_empty}
        \includegraphics[width=\linewidth]{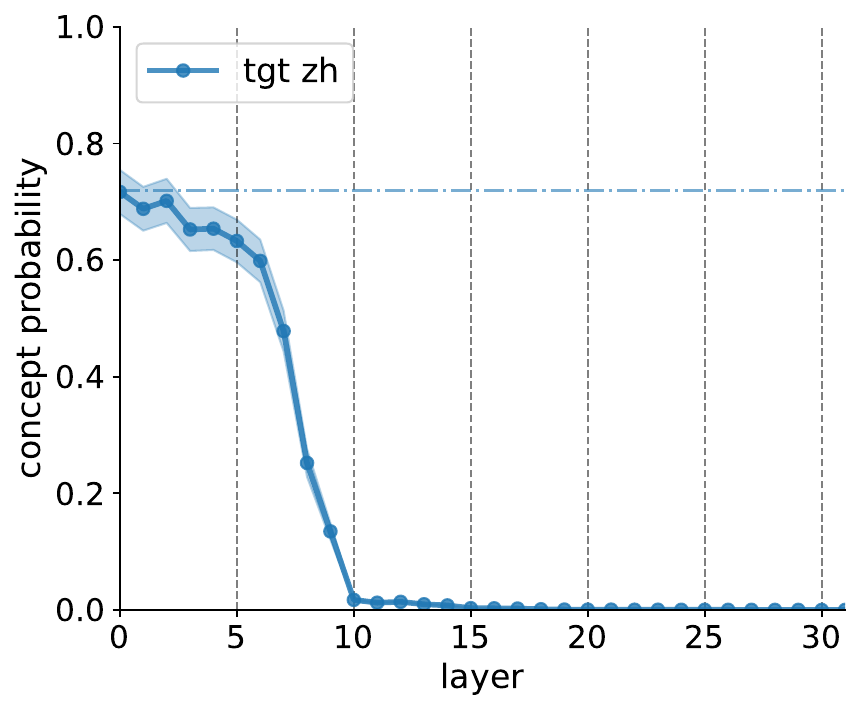}
    \end{subfigure}
    \vspace{1em}
    \begin{subfigure}[t]{0.49\linewidth}
        \caption{Random prompt with ``@'' instead of quotation mark}
        \label{fig:rand_at}
        \includegraphics[width=\linewidth]{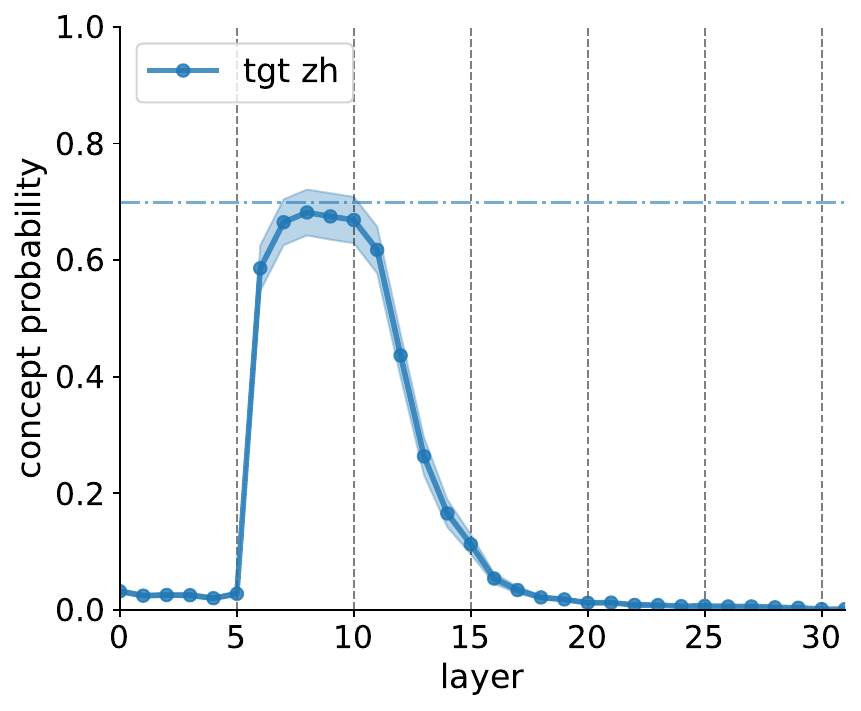}
    \end{subfigure}
    \hfill
    \begin{subfigure}[t]{0.49\linewidth}
    \centering
        \caption{Random shuffled prompt (random hidden state)}
        \label{fig:random_shuffled}
        \includegraphics[width=\linewidth]{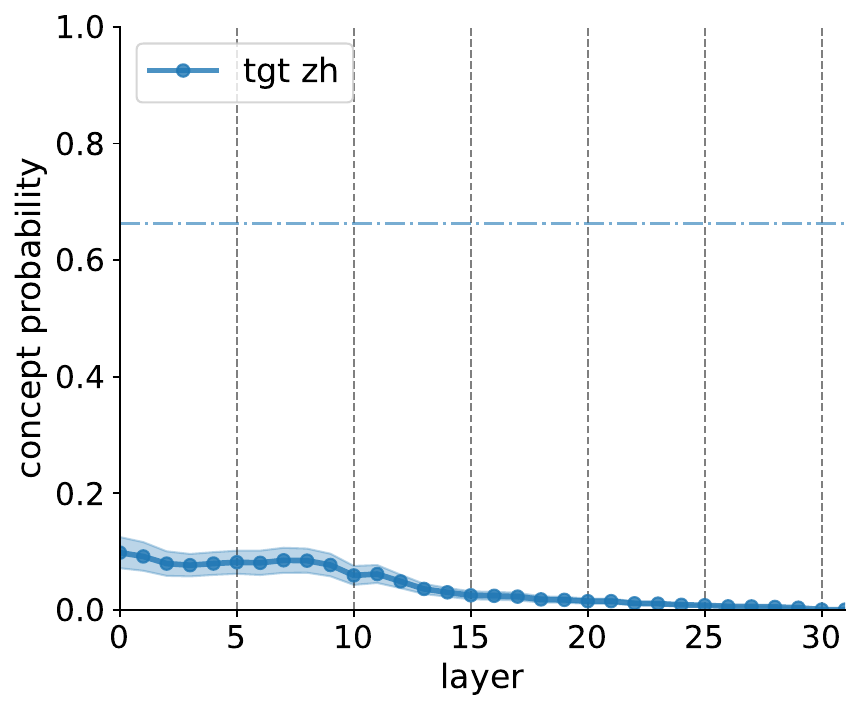}
    \end{subfigure}
    \caption{(a) activation patching experiment with a randomized source prompt (random concepts, and languages, but same template) and a \cpt{FR} to \cpt{ZH} target prompt. (b) we construct a source prompt with empty context. (c) we replace the quotation mark with @ in the random source prompt from (a). (d) we randomly shuffle the source prompts from (c). We patch at the last token respectively. For each of the plots, the x-axis shows at which layer the patching was performed during the forward pass on the target prompt and the y-axis shows the probability of predicting the correct concept in language $\ell$ (see legend). We only plot the target (``tgt'') concept, as there is no source concept to predict. We report means and 95\% Gaussian confidence intervals computed over 200 source-, target prompt pairs.}
    \label{fig:randomandact}
\end{figure}

\xhdr{Same template} In Figure~\ref{fig:random}, we randomized both input and output language as well as concepts in the source prompts, resulting in prompts of the following form:
\prompt{
{\color{red}A}: ``$\text{\cpt{CAT}}^\text{\cpt{DE}}$'' - {\color{red}B}: ``$\text{{\color{red}\cpt{DOG}}}^\text{\cpt{IT}}$''

{\color{red}A}: ``$\text{\cpt{OWL}}^\text{\color{red}\cpt{JA}}$'' - {\color{red}B}: ``$\text{{\color{red}\cpt{SUN}}}^\text{\color{red}\cpt{HI}}$''

{\color{red}A}: ``$\text{\cpt{ICE}}^\text{\color{red}\cpt{FR}}$'' - {\color{red}B}: ``
}

By doing this, the latent of the source prompt is similar in terms of prompt structure, but the model cannot infer a task vector specifying the output language since the source prompt instantiates an impossible task (to predict a random word in a random language). As shown in Figure~\ref{fig:random}, for layers 0--11, we observe no drop in the accuracy, which confirms our hypothesis that in those layers the latent at last token position contains no information specific to the translation task. 

Instead, we think that in our chosen prompting template the last token, which is a quotation mark, merely indicates \emph{where to put the translation result}. In order to investigate this, we performed further patching experiments investigating how changes in the prompting template in the source prompt affects the target forward pass ability to compute an answer. 

\xhdr{Empty context} For example, replacing the source prompt with an empty prompt, merely containing \miniprompt{B: ``} results in Figure~\ref{fig:random_empty}. In contrast to Figure~\ref{fig:random}, the target concept in target language probability drops already starting from layer 4. We think this is due to the fact that until layer 4 the quotation mark token information which is shared among the two prompting templates ``dominates'' the latent representation and is not yet converted to a task specific position marker yet. Then, starting from layer 4 the latent representation of the last token also aggregates task specific information, in particular, the fact that the quotation mark in this task actually marks the position after which the translated word should be decoded. As a result, replacing the task specific quotation mark embedding, which contains the information that the translated word comes next, with the ``empty-context''-one, which does not contain this information, results in a performance drop.

\xhdr{Modified template} Next, replacing the quotation marks by ``@'' (Figure~\ref{fig:rand_at}) in the random prompt, i.e.,
\prompt{
{\color{red}A}: @$\text{\cpt{CAT}}^\text{\cpt{DE}}$@ - {\color{red}B}: @$\text{{\color{red}\cpt{DOG}}}^\text{\cpt{IT}}$@

{\color{red}A}: @$\text{\cpt{OWL}}^\text{\color{red}\cpt{JA}}$@ - {\color{red}B}: @$\text{{\color{red}\cpt{SUN}}}^\text{\color{red}\cpt{HI}}$@

{\color{red}A}: @$\text{\cpt{ICE}}^\text{\color{red}\cpt{FR}}$@ - {\color{red}B}: @
}

leads to a drop of performance for early layers, but for layers 5--11, the model is not much affected by the patching. We postulate that at those layers, position-marker tokens have been already mapped to a general position-marker feature that is similar in between source and target forward pass, even though at input level different symbols have been used. 

\xhdr{Shuffled tokens} Lastly, in Figure~\ref{fig:random_shuffled} we try to destroy all of the shared structure in between the source and the target prompt by randomly shuffling the characters of the source prompts from the \textbf{modified template} task. As expected, the probability of the target concept in target language becomes very low (albeit surprisingly not zero), which shows that the task cannot be solved without the position marker feature.

\section{Other models and languages}\label{app:other-models}

In this section, we report results for additional models, namely, Mistral 7B~\cite{jiang2023mistral}, Llama 3 8B~\cite{dubey2024llama}, Qwen 1.5 7B~\cite{bai2023qwentechnicalreport} and Llama 2 70B~\cite{touvron2023llama}. We also include Aya 23 8B~\cite{aryabumi2024aya23openweight} for the mean patching experiment in App~\ref{app:mainexp}.

\subsection{Exploratory analysis}

\begin{figure*}[ht]
    \centering
    \begin{subfigure}[t]{0.48\linewidth}
        \caption{Mistral-7B v0.3}
        \includegraphics[width=\linewidth]{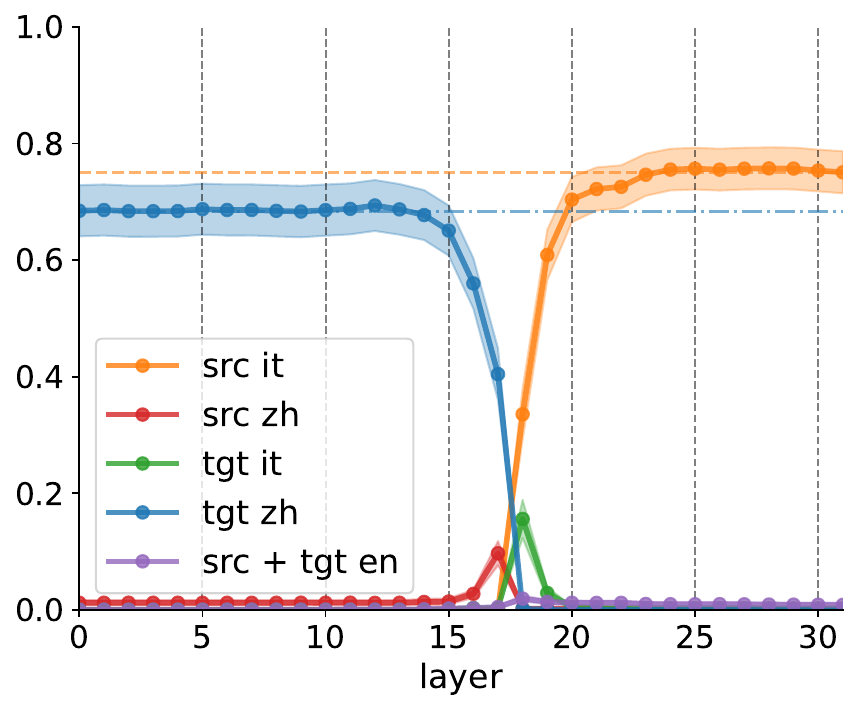}
    \end{subfigure}
    \hfill
    \begin{subfigure}[t]{0.48\linewidth}
        \caption{Llama3-8b}
        \includegraphics[width=\linewidth]{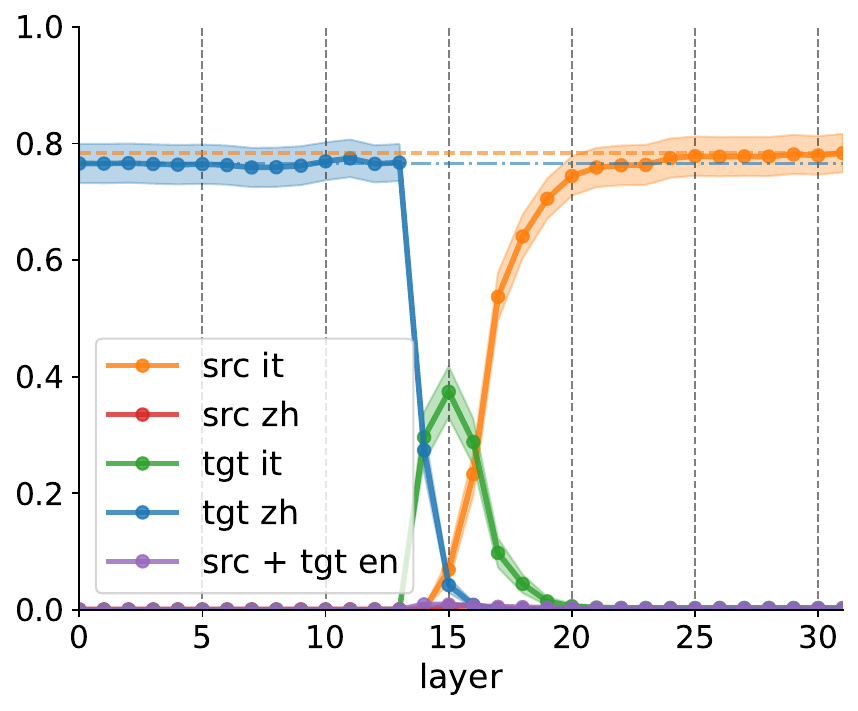}
    \end{subfigure}
    
    \vspace{1em}
    
    \begin{subfigure}[t]{0.48\linewidth}
        \caption{Qwen1.5-7B}
        \includegraphics[width=\linewidth]{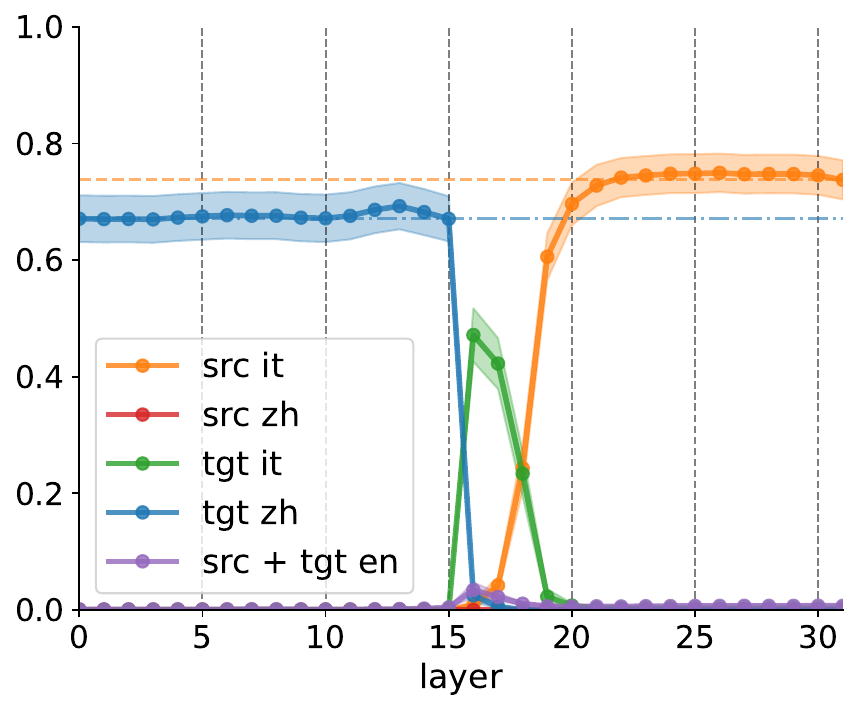}
    \end{subfigure}
    \hfill
    \begin{subfigure}[t]{0.48\linewidth}
        \caption{Llama2-70B}
        \includegraphics[width=\linewidth]{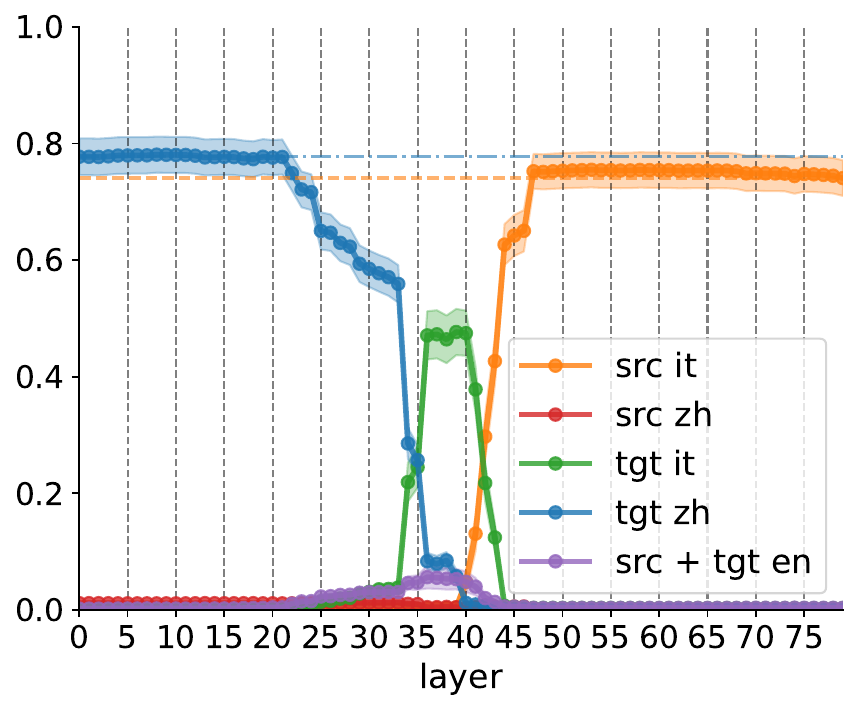}
    \end{subfigure}
    \caption{Our first patching experiment with a \cpt{DE} to \cpt{IT} source prompt and a \cpt{FR} to \cpt{ZH} target prompt with different concepts. We patch at the last token. For each of the plots the x-axis shows at which layer the patching was performed during the forward pass on the target prompt and the y-axis shows the probability of predicting the correct concept in language $\ell$ (see legend). In the legend the prefix ``src'' stands for source and ``tgt'' for target concept. The orange dashed line and blue dash-dotted line correspond to the mean accuracy on source and target prompt. We report means and 95\% Gaussian confidence intervals computed over 200 source-, target prompt pairs featuring 41 source concepts and 38 target concepts.}
    \label{fig:patchscopeallmodels}
\end{figure*}

The results of the exploratory analysis outlined in Sec.~\ref{sec:exploratory} are in Figure~\ref{fig:patchscopeallmodels}.

As can be seen in Figure~\ref{fig:patchscopeallmodels}, the target concept in source language spike is smaller for Llama 3, Mistral 7B v0.3 and Qwen 1.5 7B. This hints that for those models, $z_{\ell^{(\text{out})}}$ and $C$ computation overlap more than for Llama-2-7B.

\subsection{Ruling out hypotheses}
\label{app:mainexp}
In this section, we report results for the experiments performed in Sec.~\ref{sec:ruling-out}.

\begin{figure*}[ht]
    \centering
    \begin{minipage}{\textwidth}
        \centering
        \begin{tabular}{@{}ccc@{}}
            (a) \textbf{Single source setup} & (b) \textbf{Mean over contexts} & (c) \textbf{Mean over language pairs} \\[1pt]
            
            \includegraphics[width=0.25\textwidth]{figures/plots/obj_patching.pdf} &
            \includegraphics[width=0.25\textwidth]{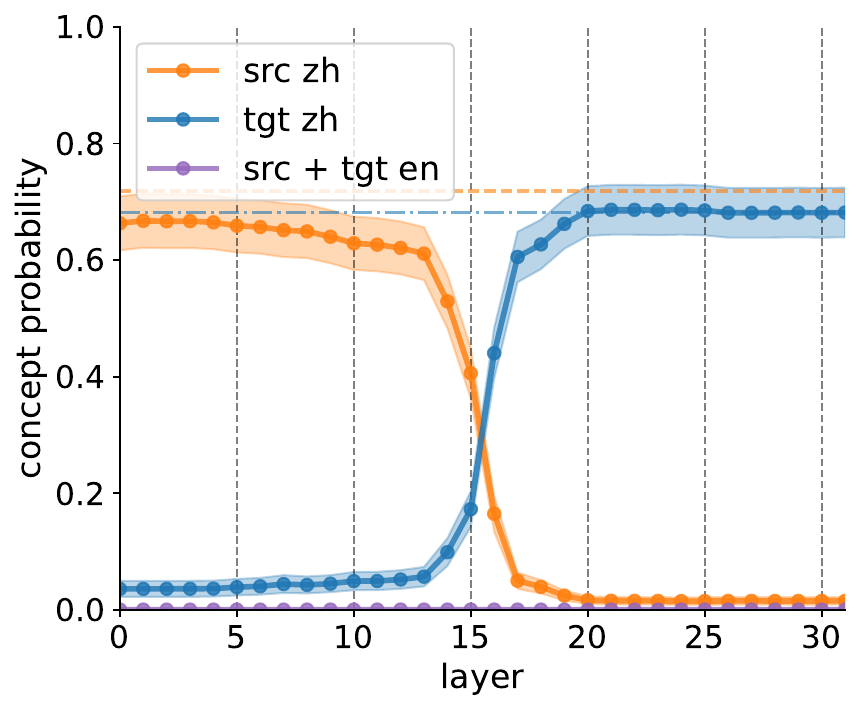} &
            \includegraphics[width=0.25\textwidth]{figures/plots/mean_obj_patching.pdf} \\
            \multicolumn{3}{c}{Llama-2 7B} \\[3pt]
            
            \includegraphics[width=0.25\textwidth]{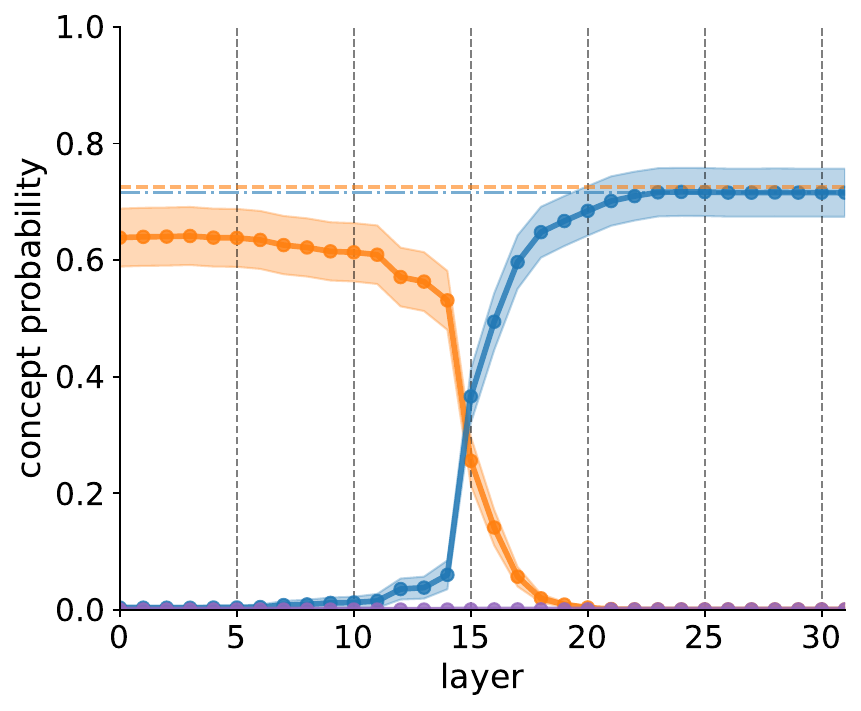} &
            \includegraphics[width=0.25\textwidth]{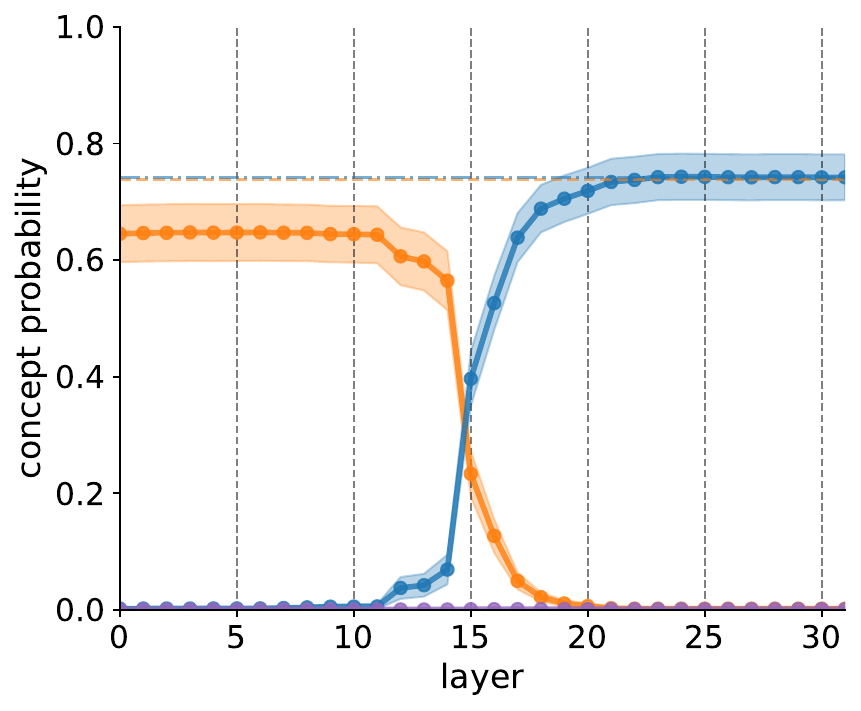} &
            \includegraphics[width=0.25\textwidth]{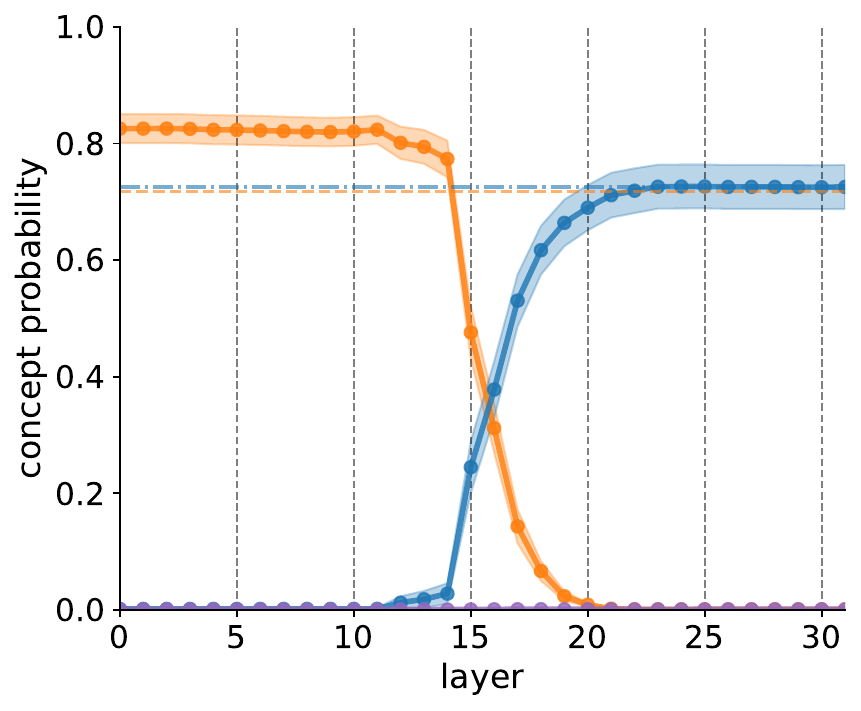} \\
            \multicolumn{3}{c}{Llama-3 8B} \\[3pt]

            \includegraphics[width=0.25\textwidth]{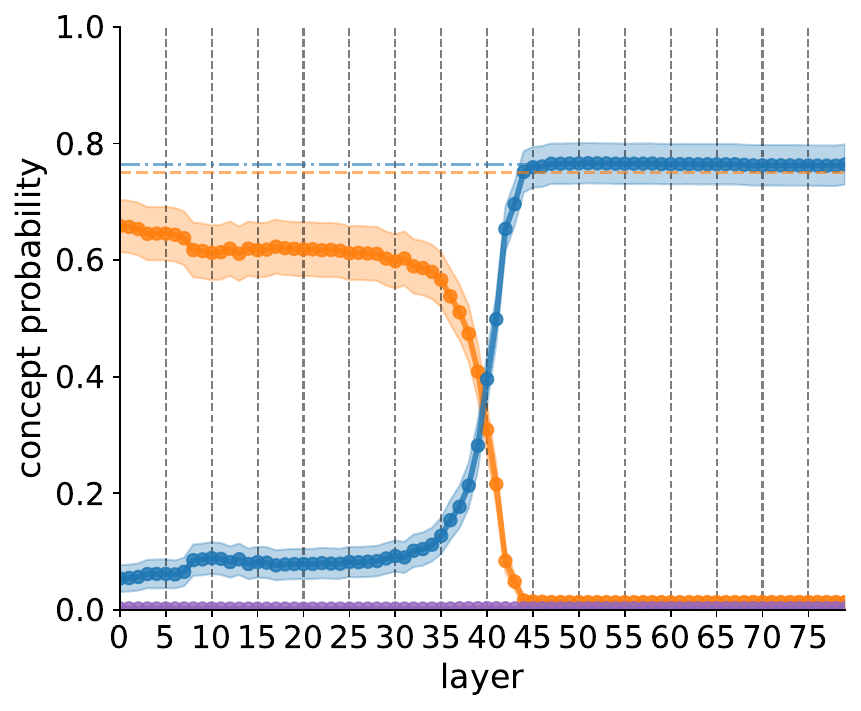} &
            \includegraphics[width=0.25\textwidth]{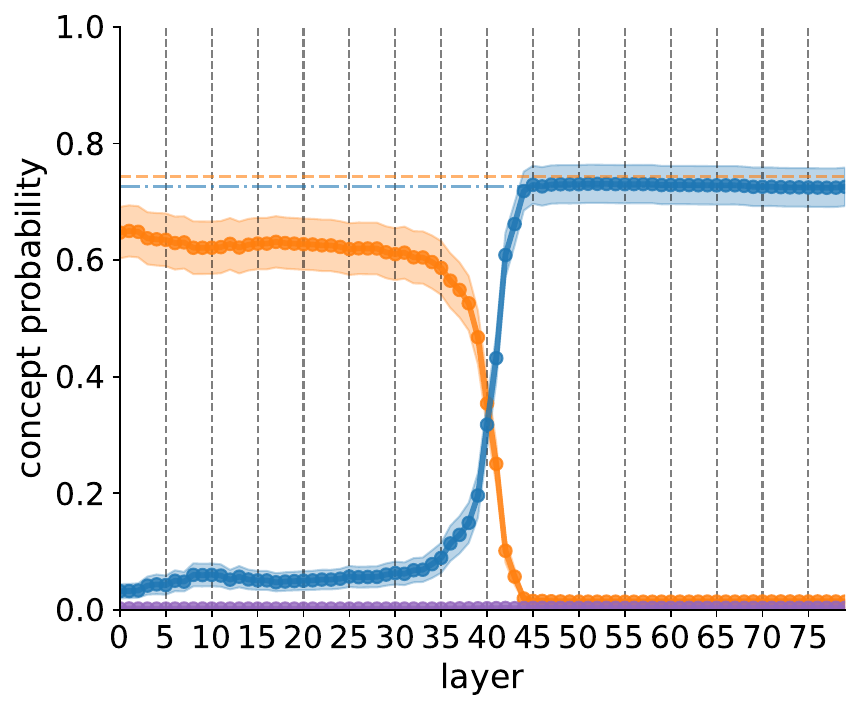} &
            \includegraphics[width=0.25\textwidth]{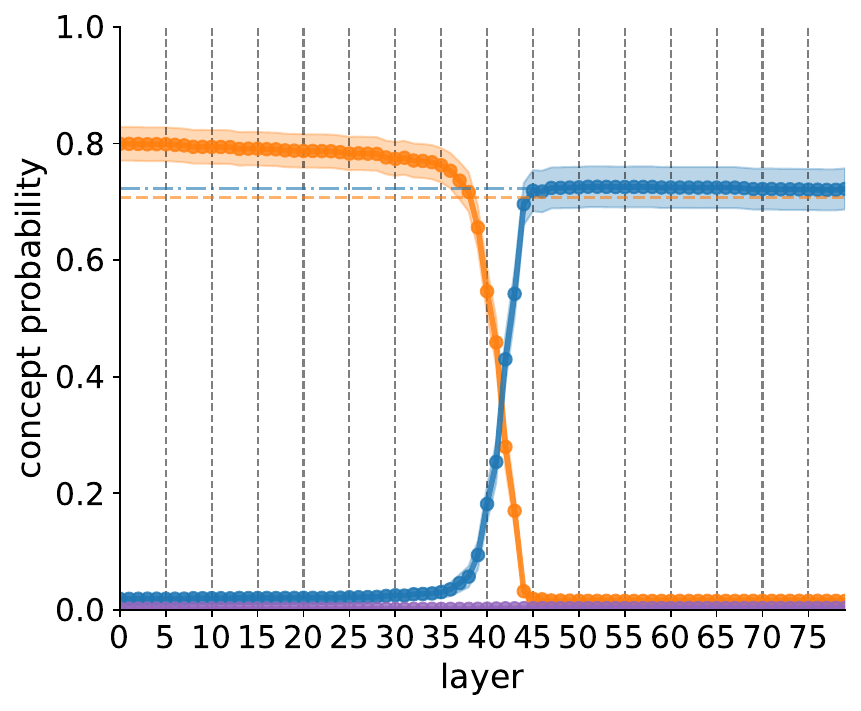} \\
            \multicolumn{3}{c}{LLama-2 70B}
        \end{tabular}
    \end{minipage}
    \caption{Here we use different input languages (\cpt{DE}, \cpt{FR}), different concepts, different output languages (\cpt{IT}, \cpt{ZH}) in (a). In (b) we use the same source and target language pairs as in (a). In (c) we use multiple source input languages \cpt{DE}, \cpt{NL}, \cpt{ZH}, \cpt{ES}, \cpt{RU} and output languages \cpt{IT}, \cpt{FI}, \cpt{ES}, \cpt{RU}, \cpt{KO}. We patch at the last token of the concept-word at all layers from $j$ to $31$. In (a) we patch latents from the single source prompt. In (b) for each concept, we patch the average latent over different few-shot \cpt{DE} to \cpt{IT} translation contexts. In (c) we patch the mean of the latents over the source prompts. 
    For each of the plots, the x-axis shows at which layer the patching was performed during the forward pass on the target prompt and the y-axis shows the probability of predicting the correct concept in language $\ell$ (see legend). The prefix ``src'' stands for source and ``tgt'' for target concept. We report means and 95\% Gaussian confidence intervals computed over a dataset of size 200.}
    \label{fig:obj_patching_full}
\end{figure*}

\begin{figure*}[ht]
    \centering
    \begin{minipage}{\textwidth}
        \centering
        \begin{tabular}{@{}ccc@{}}
            (a) \textbf{Single source setup} & (b) \textbf{Mean over contexts} & (c) \textbf{Mean over language pairs} \\[1pt]

            \includegraphics[width=0.25\textwidth]{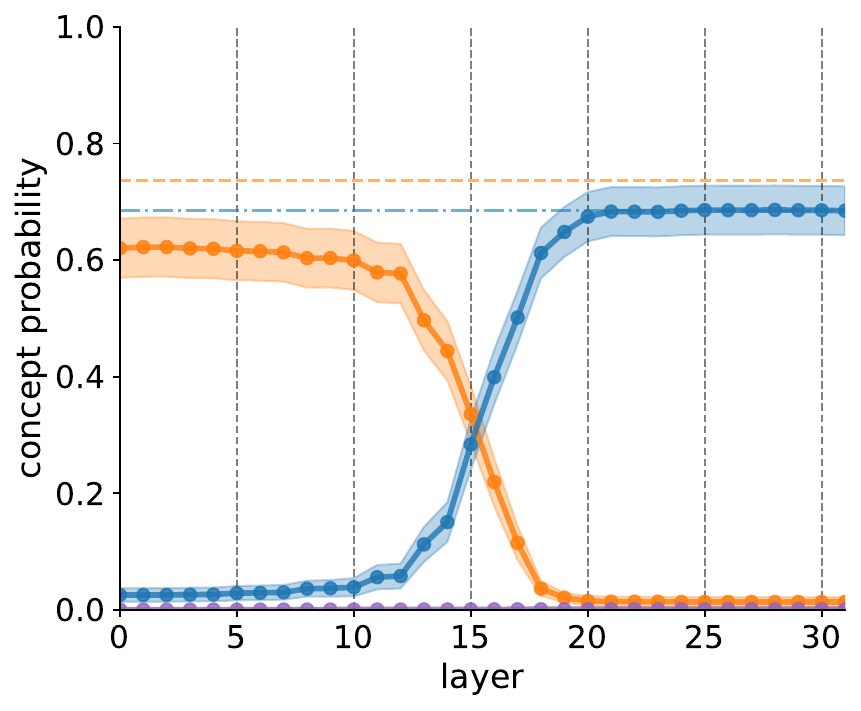} &
            \includegraphics[width=0.25\textwidth]{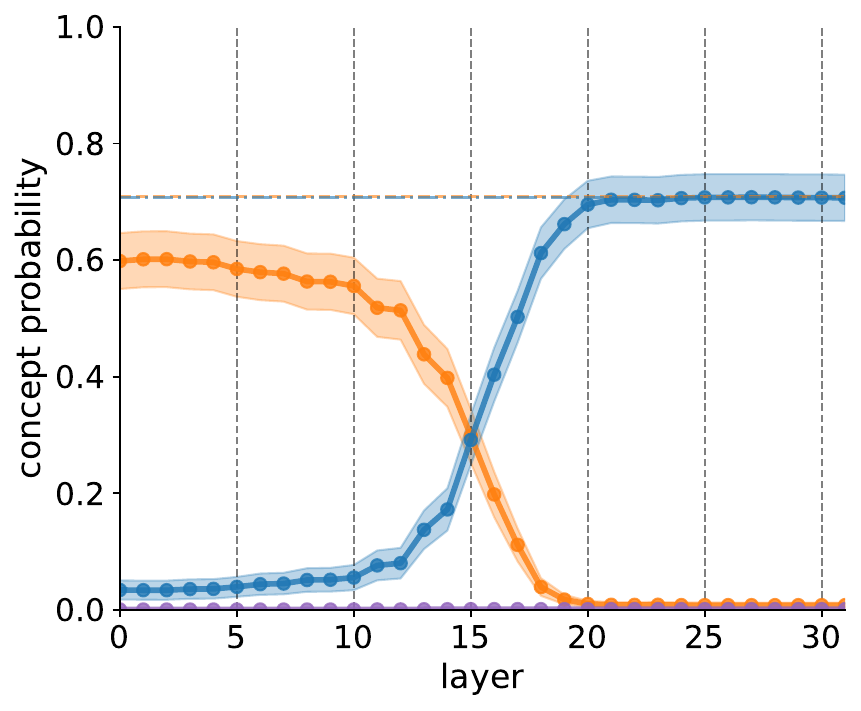} &
            \includegraphics[width=0.25\textwidth]{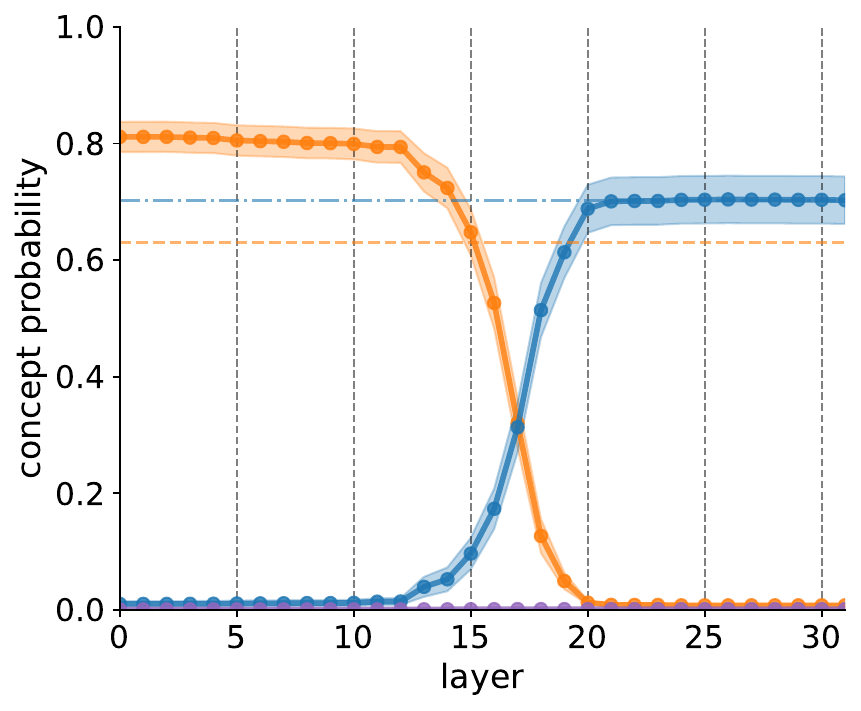} \\
            \multicolumn{3}{c}{Mistral 7B v0.3}\\[3pt]
            
            \includegraphics[width=0.25\textwidth]{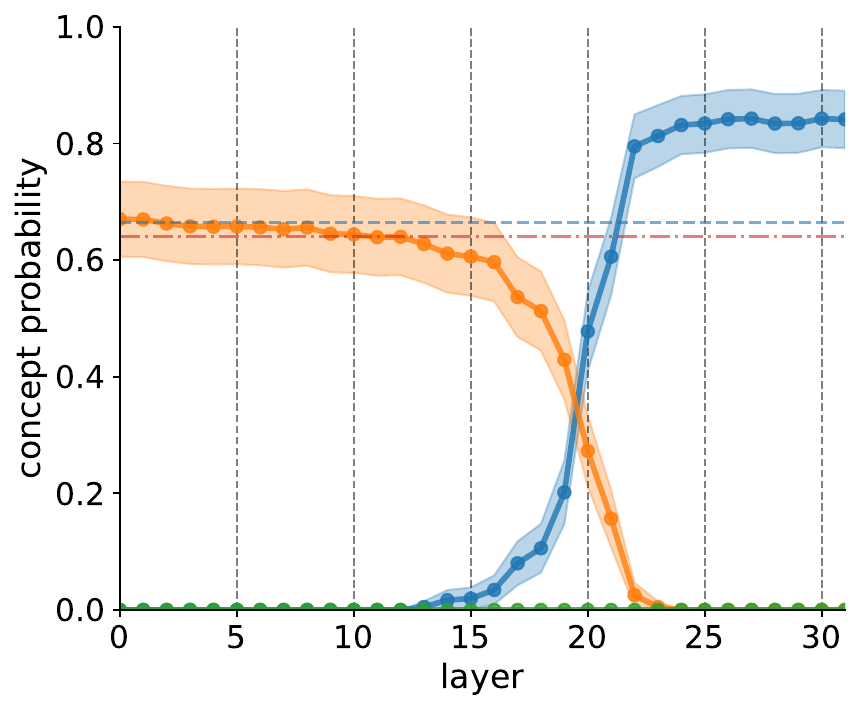} &
            \includegraphics[width=0.25\textwidth]{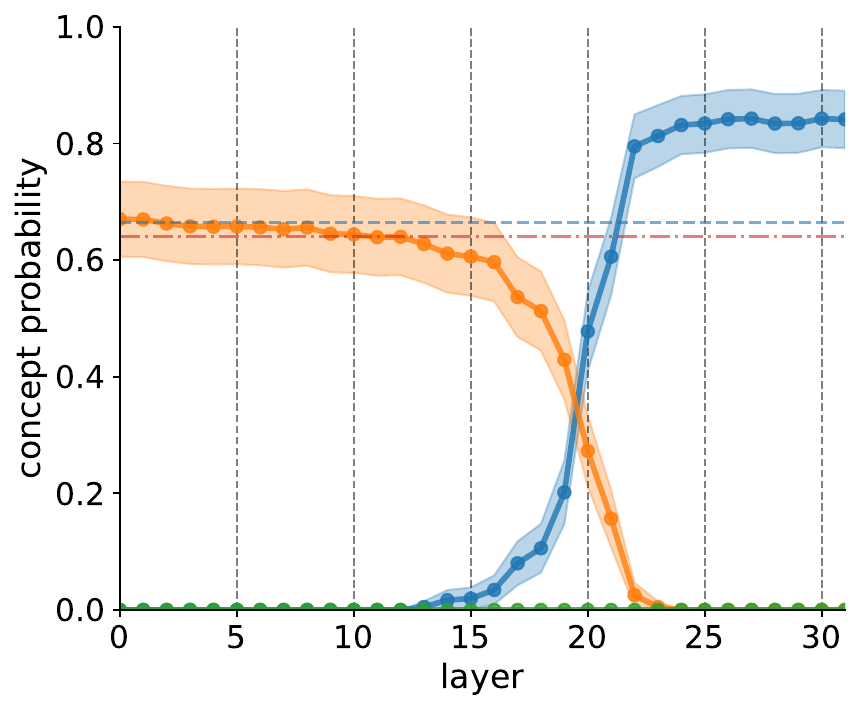} &
            \includegraphics[width=0.25\textwidth]{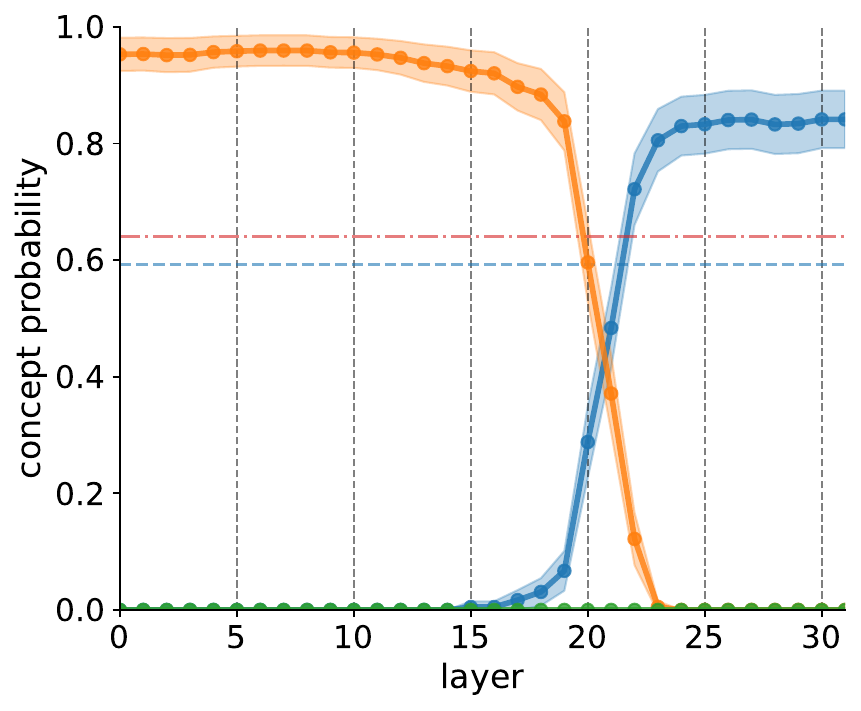} \\
            \multicolumn{3}{c}{Aya 23 8B}\\[3pt]
            
            \includegraphics[width=0.25\textwidth]{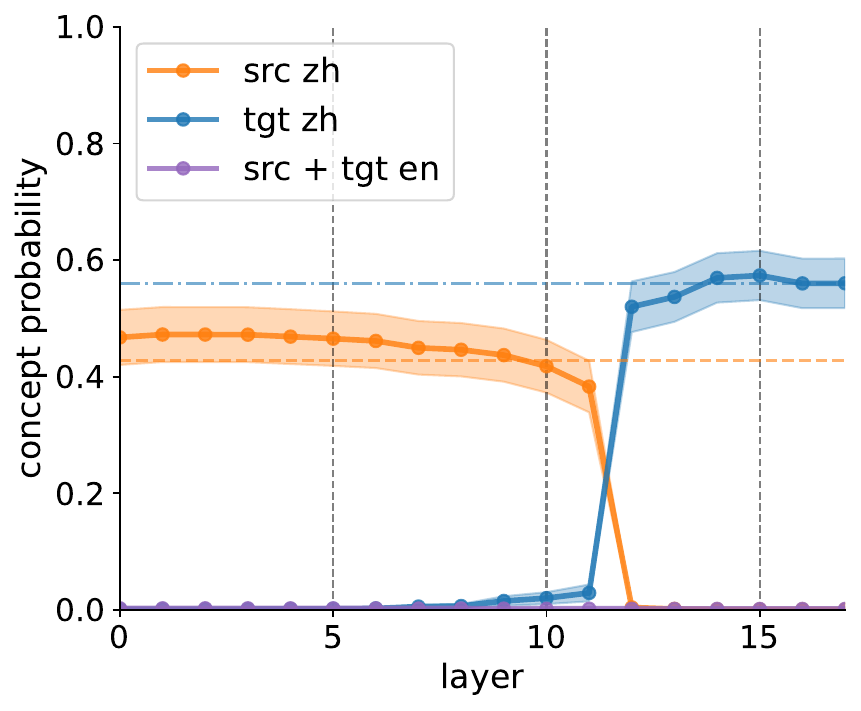} &
            \includegraphics[width=0.25\textwidth]{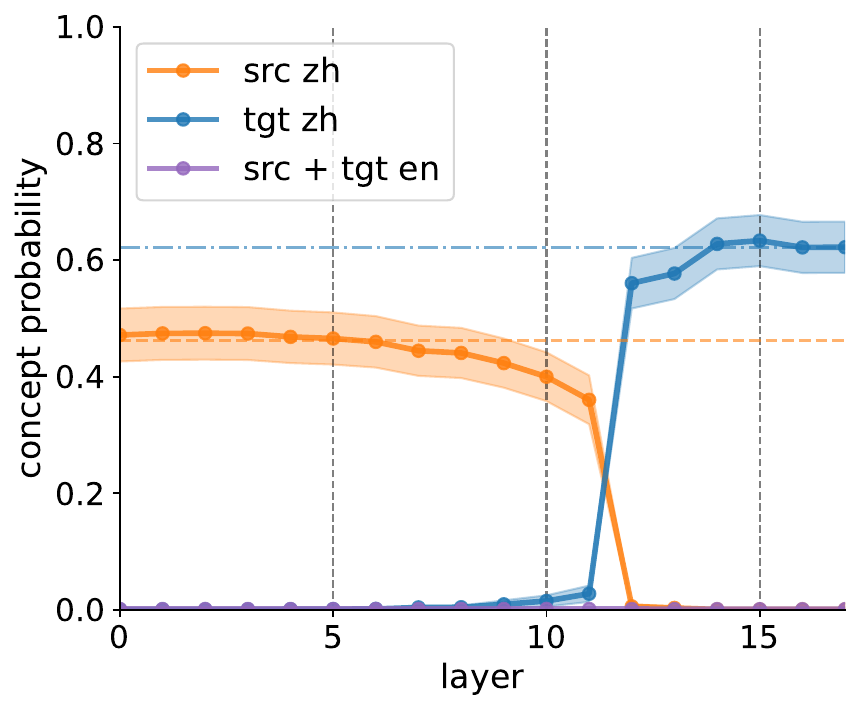} &
            \includegraphics[width=0.25\textwidth]{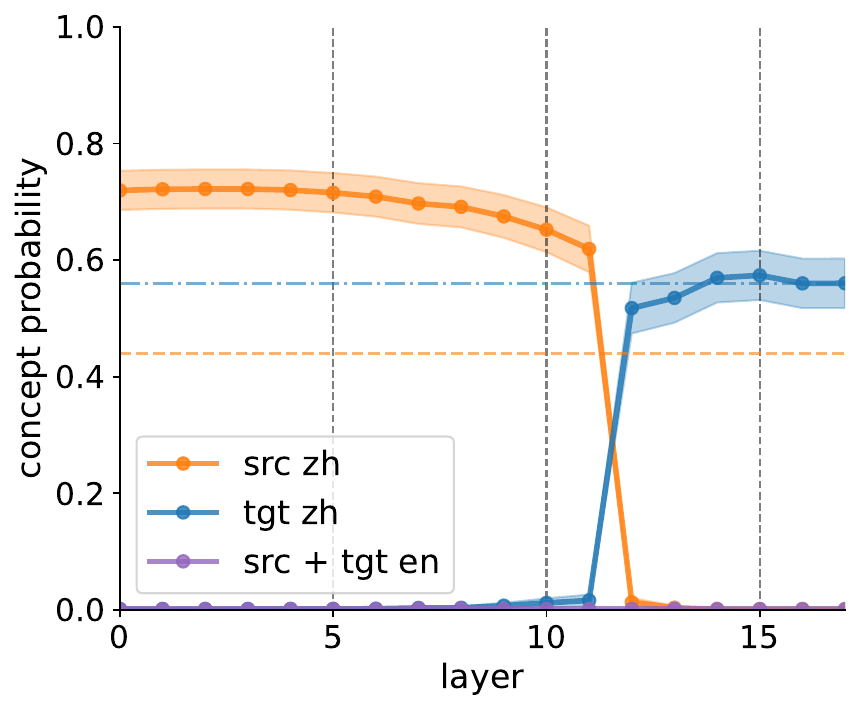} \\
            \multicolumn{3}{c}{Gemma 2B}
        \end{tabular}
    \end{minipage}
    \caption{Here we use different input languages (\cpt{DE}, \cpt{FR}), different concepts, different output languages (\cpt{IT}, \cpt{ZH}) in (a). In (b) we use the same source and target language pairs as in (a). In (c) we use multiple source input languages \cpt{DE}, \cpt{NL}, \cpt{ZH}, \cpt{ES}, \cpt{RU} and output languages \cpt{IT}, \cpt{FI}, \cpt{ES}, \cpt{RU}, \cpt{KO}. We patch at the last token of the concept-word at all layers from $j$ to $31$. In (a) we patch latents from the single source prompt. In (b) for each concept, we patch the average latent over different few-shot \cpt{DE} to \cpt{IT} translation contexts. In (c) we patch the mean of the latents over the source prompts. 
    For each of the plots, the x-axis shows at which layer the patching was performed during the forward pass on the target prompt and the y-axis shows the probability of predicting the correct concept in language $\ell$ (see legend). The prefix ``src'' stands for source and ``tgt'' for target concept. We report means and 95\% Gaussian confidence intervals computed over a dataset of size 200.}
    \label{fig:obj_patching_full2}
\end{figure*}

In addition, instead of just patching in the mean over different language pairs (Figure~\ref{fig:obj_patching_full}c, \ref{fig:obj_patching_full2}c), we also patch in the mean over contexts composed of different concept words in Figure~\ref{fig:obj_patching_full}b, \ref{fig:obj_patching_full2}b. In particular, we take the mean over 5 different few-shot contexts from the same language pair. E.g.:
\noindent
\prompt{Deutsch: ``Dorf'' - Italiano: ``villaggio''\\
 \vdots\\
Deutsch: ``Buch
}\linebreak
\vskip-0.5\baselineskip
\adjustbox{raise=1\baselineskip}{\vdots}\\
\prompt{
Deutsch: ``Zitrone'' - Italiano: ``limone''\\
 \vdots\\
Deutsch: ``Buch
}

Our results in Figure~\ref{fig:obj_patching_full} and Figure~\ref{fig:obj_patching_full2} show that the mean over contexts does not increase $P(C^{\ell_T}_S)$, whereas the mean over language pairs does. This is intuitive, since there may be some languages in which the mapping from words to concept features results in the correct concept feature vector. Therefore, averaging over different language pairs can increase the signal about the source concept. However, having additional random contexts stemming from the same language pair does not bring in any information about the source concept. 

\textbf{Note that Figure~\ref{fig:patchscopeallmodels}, Figure~\ref{fig:obj_patching_full} and Figure~\ref{fig:obj_patching_full2}  are on the next two pages.} 

\begin{figure*}
    \centering
    \begin{minipage}[t]{0.5\textwidth}
        \centering
        \includegraphics[width=\linewidth]{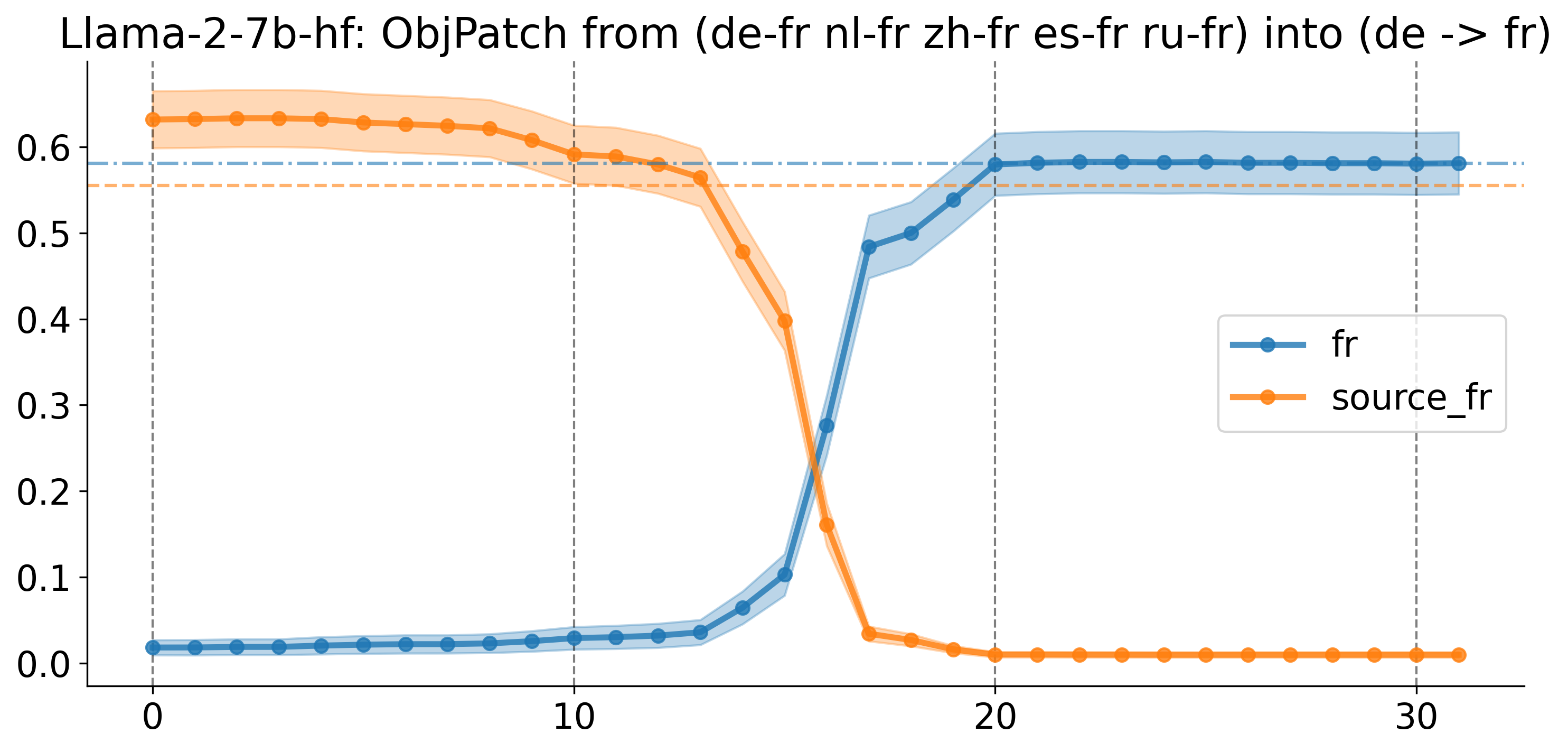}
        \label{fig:fr}
    \end{minipage}%
    \begin{minipage}[t]{0.5\textwidth}
        \centering
        \includegraphics[width=\linewidth]{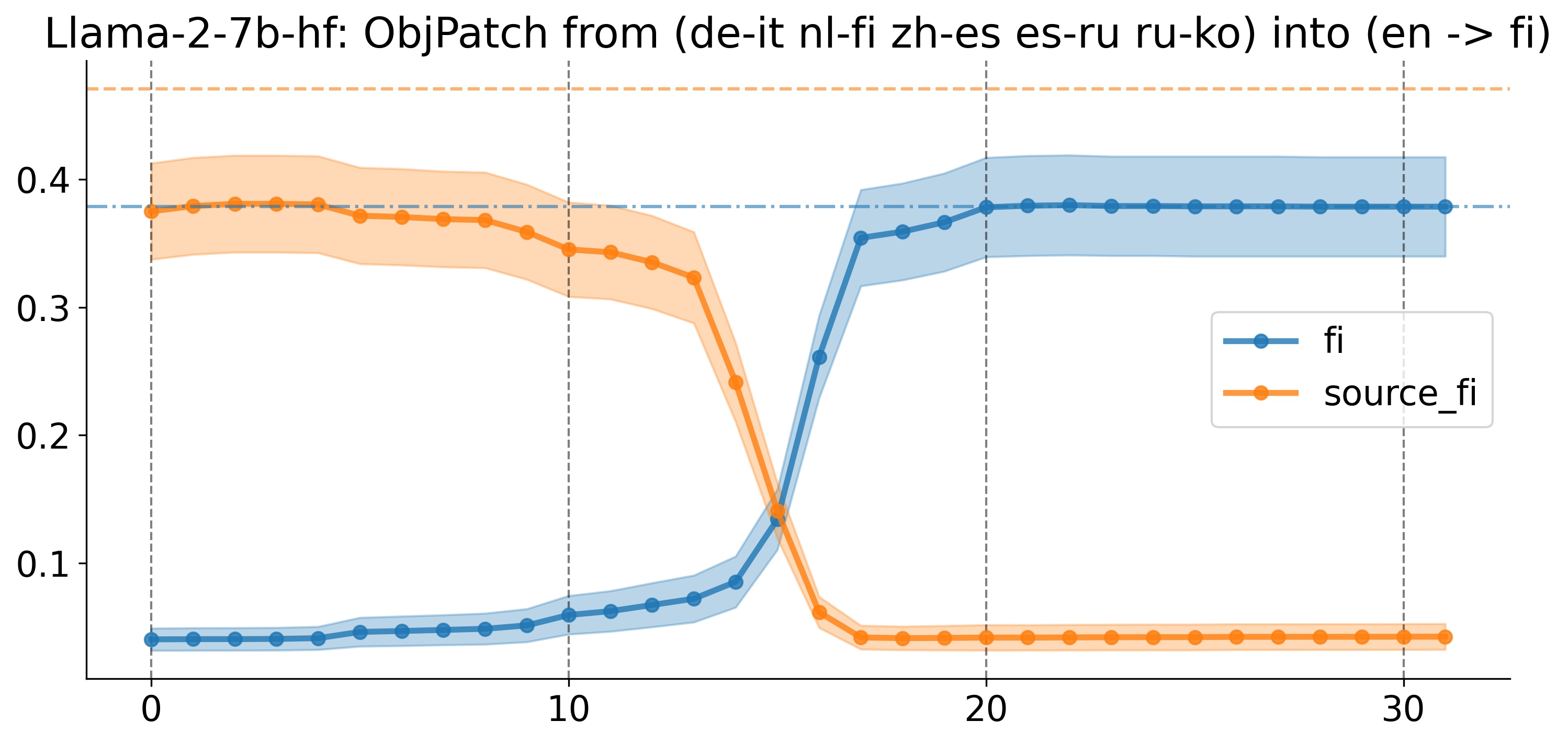}
        \label{fig:plot3}
    \end{minipage}
    \begin{minipage}[t]{0.5\textwidth}
        \centering
        \includegraphics[width=\linewidth]{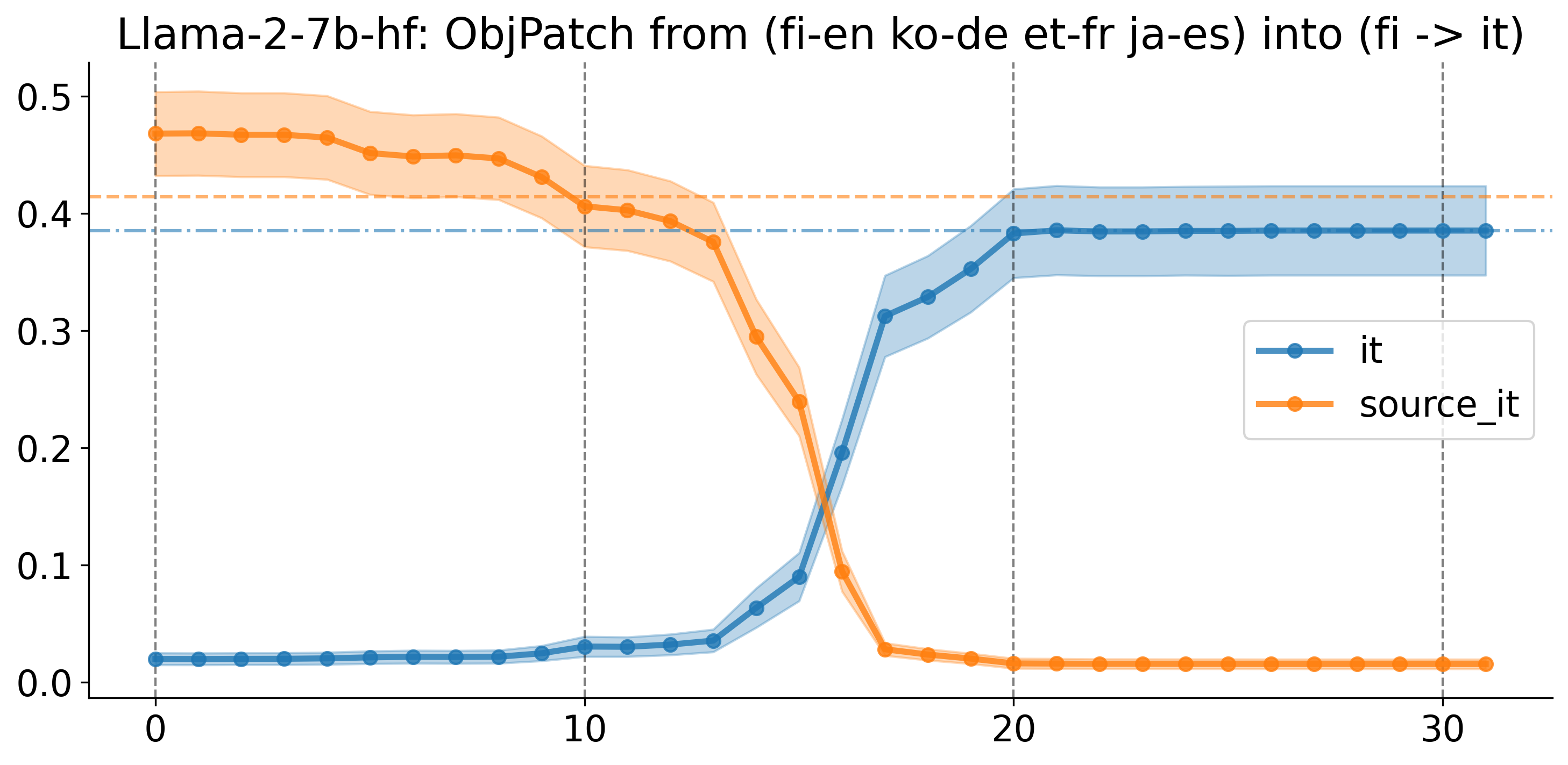}
        \label{fig:it}
    \end{minipage}%
    \begin{minipage}[t]{0.5\textwidth}
        \centering
        \includegraphics[width=\linewidth]{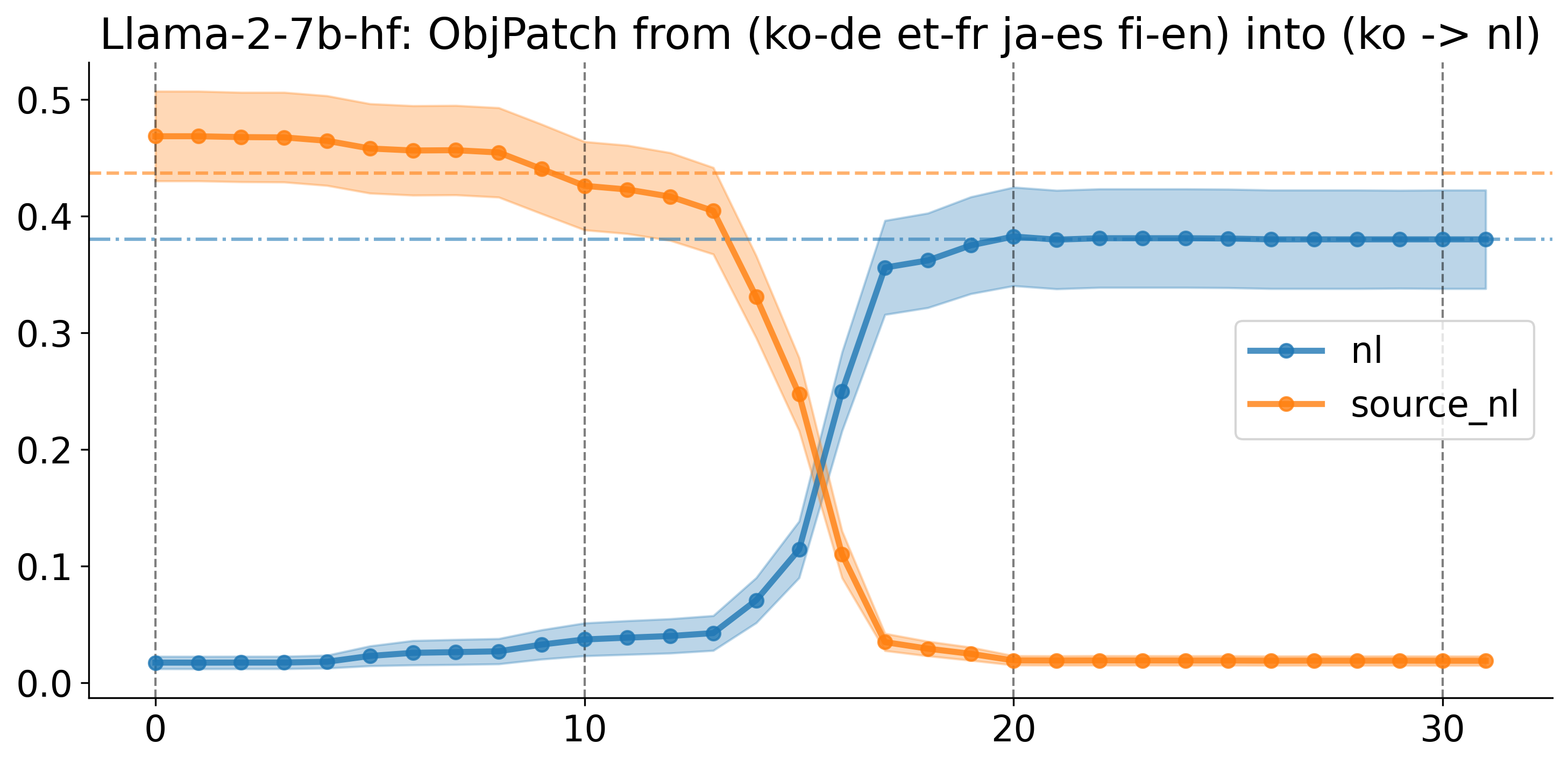}
        \label{fig:nl}
    \end{minipage}
    \caption{Mean patching experiment replicated on multiple languages with 200 pairs each and 95\% confidence interval.}
    \label{fig:other_langs}
\end{figure*}

\subsection{Similarity comparison}
\label{app:similarities}
\begin{figure*}[ht]
    \centering
    \begin{minipage}{\textwidth}
        \centering
        \begin{tabular}{@{}c@{}}
            \includegraphics[width=\textwidth]{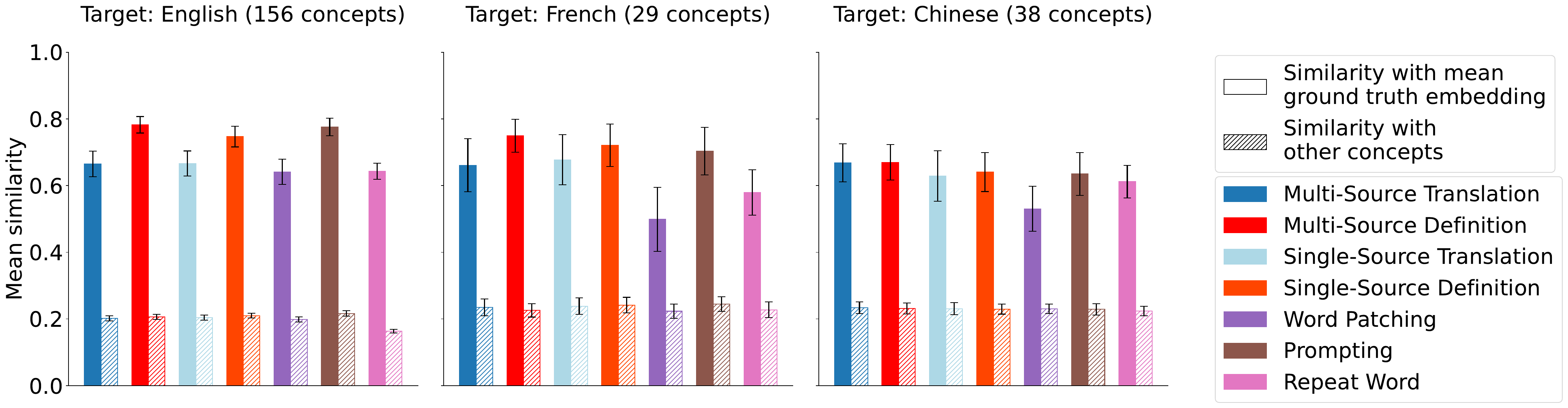} \\
            (a) Aya 23 8B \\[3pt]
            \includegraphics[width=\textwidth]{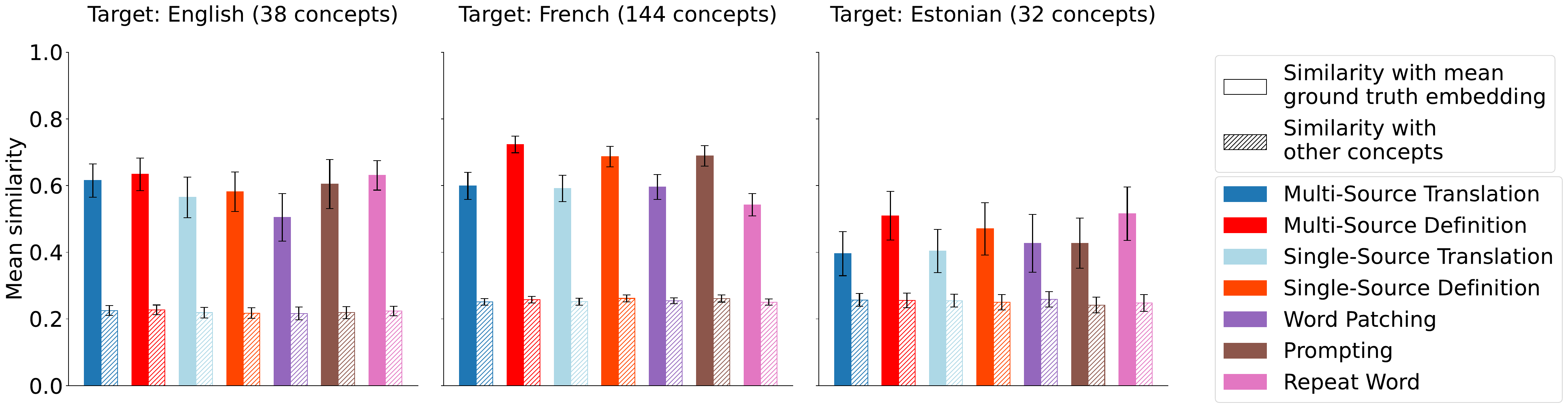} \\
            (b) Gemma 2 2B
        \end{tabular}
    \end{minipage}
    \caption{Mean similarity between the definition and the mean embedding of the ground truth definitions, as well as the mean similarity between the definition embedding and the embeddings of the definitions of the other concepts in the dataset. For Aya, the results are presented for three target languages: English (with source languages French and German and input language Spanish), French (with source languages Korean, Japanese, Estonian, Finnish and input language English), and Chinese (with source languages Italian, Finnish, Spanish, Russian, Korean and input languages German, Dutch, Chinese, Spanish, Russian). For Gemma, we show English (with source languages Italian, Finnish, Spanish, Russian, Korean and input languages German, Dutch, Chinese, Spanish, Russian), French (with source languages Spanish, German and input language Italian), and Estonian (with source languages English, French, Chinese, German and input language Hindi). We report means and 95\% Gaussian confidence intervals computed over the dataset.}
    \label{fig:other_similarities}
\end{figure*}

\begin{figure*}
\includegraphics[width=\textwidth]{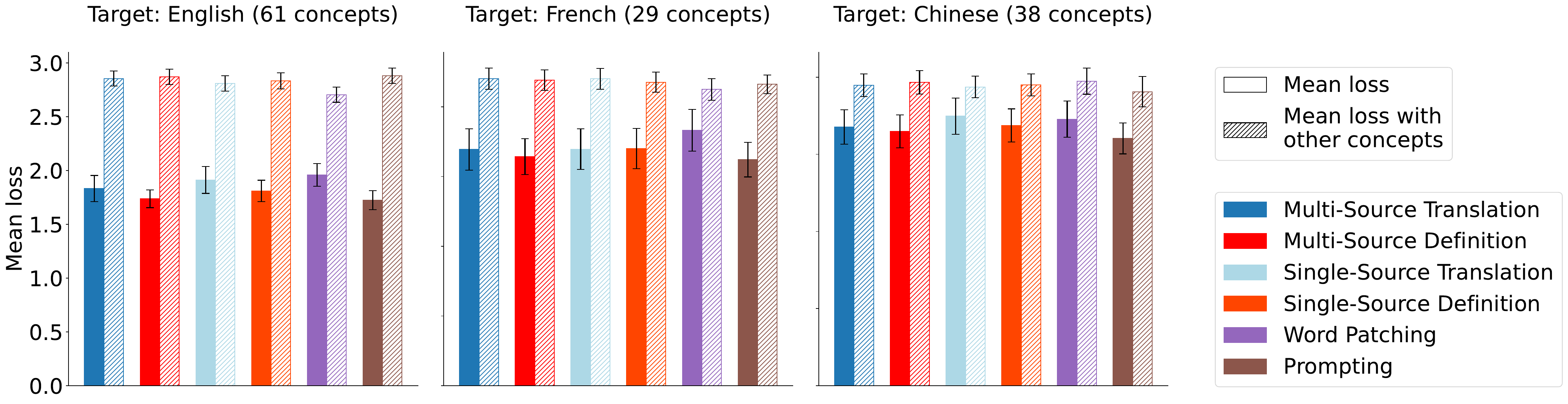}
    \caption{Mean loss on the ground truth definitions. We report 95\% confidence intervals.}
    \label{fig:loss}
\end{figure*}

We experiment with other languages and models in Figure~\ref{fig:other_similarities} and get the same trends as with our results in Figure~\ref{fig:def_sims}. We also provide results for another experiment in which instead of measuring embedding similarities, we measure perplexity on ground truth definitions in Figure~\ref{fig:loss}. We did not include this metric in our analysis as it seemed to have less granularity and is more sensitive to the syntax rather than being focused on the semantic.

\end{document}